\documentclass[preprint,12pt]{elsarticle}
\usepackage[pdfborder={0 0 0}]{hyperref}
\usepackage[margin=1in]{geometry}

\biboptions{numbers,sort&compress}
\usepackage{amssymb}
\usepackage{amsmath}
\usepackage{xcolor}

\usepackage{comment}
\usepackage{multirow,booktabs}
\usepackage{fixme}
\usepackage{amsmath}
\usepackage{subcaption}
\usepackage{makecell}
\usepackage{amsfonts} 
\usepackage{hyperref}
\usepackage{booktabs}
\usepackage{xcolor}
\usepackage{lipsum}
\usepackage{verbatim}
\usepackage{array}
\usepackage[shortcuts]{extdash}
\usepackage{xcolor}
\usepackage{hyperref}

\usepackage{microtype}
\usepackage{xurl}
\hypersetup{
    colorlinks = true,      
    linkcolor  = blue,      
    citecolor  = blue,      
    urlcolor   = blue       
}

\DeclareMathSymbol{\shortminus}{\mathbin}{AMSa}{"39}

\journal{Computers and Geotechnics}

\begin{document}

\begin{frontmatter}



\title{Temporal Extrapolation of Elastic Seismic Wave Propagation in Layered Media Using Neural Operators}


\author[1]{Dibyajyoti Nayak\fnref{fn1}} %% Author name
\author[2]{Yuze Pu\fnref{fn1}}
\author[2]{Kami Mohammadi\fnref{fn2}}
\author[1]{Somdatta Goswami\fnref{fn2}\corref{cor1}}
\ead{somdatta@jhu.edu}

\fntext[fn1]{Authors contributed equally.}
\fntext[fn2]{Authors advised equally.}
\cortext[cor1]{Corresponding author}
\affiliation[1]{organization={Department of Civil and Systems Engineering, Johns Hopkins University},%Department and Organization
            %addressline={}, 
            city={Baltimore},
            %postcode={}, 
            state={Maryland},
            country={U.S.A.}}
\affiliation[2]{organization={Department of Civil and Environmental Engineering, University of Utah},%Department and Organization
            %addressline={}, 
            city={Salt Lake City},
            %postcode={}, 
            state={Utah},
            country={U.S.A.}}

%% Abstract
\begin{abstract}
Temporal prediction of seismic velocity fields beyond the time interval represented in training remains challenging for neural operator models. This study develops a coupled neural operator framework for stress-conditioned temporal prediction in layered elastic media. A time-integrator-embedded Deep Operator Network (TI-DeepONet) advances the velocity field through an embedded fourth-order Runge--Kutta (RK4) scheme, while a Fourier Neural Operator (FNO) learns a residual correction to the predicted spatial field. The two components are trained jointly in a predictor--corrector configuration. Prediction is initialized with reference velocity fields at the first two time levels and proceeds autoregressively, with reference stress fields and prescribed excitation supplied throughout rollout, including beyond the training interval. Four numerical cases consider horizontally polarized shear (SH) waves in four-layer soil columns with fixed total depths of 50 and 280~m, excited by third-order Gaussian-derivative (DOG3) wavelets with selected central frequencies between 0.5 and 6~Hz. Layer thicknesses and shear-wave velocities vary across profiles while total depth remains fixed. Evaluations on test profiles extend beyond the training interval by up to twice its duration. The predictions reproduce principal velocity arrivals and depth-dependent response patterns, with accuracy varying across profiles. Across the four cases, mean peak-normalized root-mean-square errors (NRMSEs) range from 0.53\% to 2.31\% over the full evaluation windows and from 0.61\% to 2.82\% over the extrapolation intervals. Both metrics use each profile's peak absolute reference velocity over the full evaluation window as the normalizer. These results demonstrate temporal velocity prediction under reference-stress conditioning for the tested configurations; autonomous continuation of the coupled velocity--stress wavefield remains to be established.
\end{abstract}

%% Keywords

\end{frontmatter}

\section{Introduction}
\label{sec:intro}
Seismic waves propagating through near-surface soil deposits undergo changes in amplitude, frequency content, and duration that influence earthquake ground motions \cite{borcherdt1970Effects, aki1993Local, assimaki2008Site}. Understanding the spatiotemporal evolution of these waves is, therefore, central to geotechnical earthquake engineering. Although seismic wave propagation is inherently three-dimensional, one-dimensional models provide a useful representation of vertically propagating, horizontally polarized shear (SH) waves in laterally uniform, horizontally stratified deposits \cite{idriss1968Seismic, schnabel1972shake, huang2024Applicability}. Within this idealized setting, impedance contrasts between adjacent layers produce reflections and transmissions, while interference between wave arrivals shapes the response throughout the soil column \cite{thomson1950Transmission, haskell1953Dispersion, kennett1974Reflections, mohammadi2024One}. These interactions make layered elastic media a physically interpretable setting for examining how learned models represent the temporal evolution and spatial structure of seismic velocity fields.

Conventional approaches to one-dimensional wave propagation include frequency-domain transfer-function methods and time-domain finite-element and finite-difference schemes \cite{schnabel1972shake, tsai2016Comparison, hashash2010Recent}. In a first-order velocity-stress formulation, the momentum equation relates the rate of change of velocity to the spatial stress gradient and applied forcing, while the constitutive equation governs the evolution of stress. A complete forward simulation advances these coupled fields from specified initial conditions, material properties, and boundary inputs. 
% A related but more restricted learning problem is to predict the evolution of velocity when the corresponding stress history is provided. This formulation allows the temporal generalization of a learned velocity-update operator to be examined under prescribed stress inputs, while retaining the spatial complexity of the layered-medium response. The present study adopts this restricted setting: it evaluates recursive velocity prediction conditioned on reference stress fields, including at times beyond the interval represented in training.
A more restricted problem is to predict the velocity when the stress history is provided. In this setting the momentum equation determines the velocity rate from the supplied stress, so the task does not test whether a model can represent the coupled dynamics. It does test whether a learned update, trained on a finite time window, remains accurate and stable when applied recursively to later states whose amplitudes and spatial patterns were not seen in training. We adopt this setting as a controlled first step in evaluating the proposed framework for stress-conditioned velocity prediction.

Recent advances in machine learning have motivated the development of data-driven surrogate models for seismic response prediction. Architectures such as convolutional neural networks, recurrent neural networks, long short-term memory networks, and other deep regression models have been used to predict site amplification functions, surface ground motions, and surface-to-borehole transfer characteristics from soil profiles and input ground motions \cite{nguyen2022Machine, zhong2024Neural, choi2024Sequence, li2025Generalized}. Although such models can offer rapid evaluation after training, their accuracy may deteriorate for seismic inputs or stratigraphic configurations that are inadequately represented in the training ensemble \cite{li2025Generalized,jozinovic2022Transfer}. Physical consistency and temporal generalization also require attention, particularly when predicted states are recursively supplied as inputs for subsequent predictions \cite{moseley2020Solving,djeumou2022Neural,zhu2025Physics}.

To improve physical consistency and reduce reliance on densely sampled labeled datasets, physics-informed neural networks (PINNs) incorporate the governing partial differential equations, together with the relevant initial and boundary conditions, into the optimization objective through physics-based loss terms~\cite{RAISSI2019686}. In seismic wave-propagation applications, these physics-based constraints guide the construction of continuous spatiotemporal approximations of displacement, velocity, or stress fields \cite{ren2023Physics}. However, standard PINN formulations are generally solution-specific: a trained network approximates the response associated with a prescribed realization of material properties, geometry, excitation, and boundary conditions. Changes in these quantities commonly require retraining, fine-tuning, or an explicitly parameterized formulation. This limitation motivates learning approaches in which a single trained model can predict responses for multiple soil profiles and input motions without retraining for each configuration \cite{wang2021Learning, cho2024Parameterized}.

Neural operator learning provides a framework for predicting responses across different soil profiles and input motions using a single trained model by learning mappings between function spaces rather than approximating a single solution field \cite{lu2021learning, li2020fourier, kovachki2023neural}. Architectures such as deep operator network (DeepONet)~\cite{lu2021learning} and Fourier neural operator (FNO)~\cite{li2020fourier} have been leveraged to learn parametric partial differential equations involving variations in input functions, material coefficients, and boundary conditions \cite{kumar2025Synergistic, HAGHIGHAT2024116681, lehmann20243D}. However, extending neural operators to transient wave-propagation problems remains challenging because the underlying response is oscillatory, phase-sensitive, and strongly influenced by reflections, transmissions, and interference associated with material interfaces. During autoregressive rollout, local prediction errors can accumulate over successive time steps, producing amplitude distortion, phase drift, excessive dissipation, or instability~\cite{mccabe2023towards, lippe2023pde, michalowska2024neural, NAYAK2026118960, MANDL2026118917}. These considerations motivate evaluating both the temporal update and the spatial representation of the evolving field when predictions extend beyond the training interval. The information supplied during rollout is equally important: prediction with continuing reference stress inputs tests a different capability from autonomous evolution of the coupled velocity–stress system.

%Kami: 
%The following paragraph was removed because its main arguments - accumulated rollout errors, sensitivity to wave interactions, and the importance of temporal and spatial accurac - are already covered in the preceding paragraph. Consolidating this discussion avoids repetition and provides a more direct transition to the proposed framework.

%Extended-time temporal extrapolation remains a central challenge when applying neural operators to transient wave-propagation problems. In autoregressive rollouts, the predicted state at a given discrete time step is recursively used as the input for subsequent predictions, causing local approximation errors to propagate and accumulate over time. Over extended propagation times, these errors can distort wave amplitudes, shift arrival times, introduce phase misalignment, and eventually degrade numerical stability~\cite{mccabe2023towards, brandstetter2022message, NAYAK2026118960, MANDL2026118917}. This challenge is particularly pronounced in heterogeneous layered soil deposits, where repeated reflections, transmissions, and interference at impedance contrasts cause the wavefield to evolve through complex interactions. Under such conditions, even small phase or timing errors can progressively distort the predicted response. Reliable temporal extrapolation therefore requires an operator-learning architecture that preserves long-term temporal coherence while accurately representing the spatial structure of the evolving wavefield.

The present study introduces a framework that couples a time-integrator-embedded Deep Operator Network (TI-DeepONet)~\cite{NAYAK2026118960} with an FNO for stress-conditioned temporal prediction of seismic velocity fields. The TI-DeepONet component serves as a temporal predictor that represents the rate of change of velocity and advances the field through an embedded fourth-order Runge–Kutta scheme. The FNO acts as a residual spatial corrector that refines the predicted velocity field. The two components are trained jointly in a predictor–corrector configuration, allowing the temporal predictor and spatial corrector to co-adapt during optimization. For each test profile, prediction is initialized using reference velocity fields at the first two time levels and subsequently proceeds autoregressively. Reference stress fields at the current and preceding time levels, together with the prescribed excitation, remain inputs throughout rollout, including beyond the training interval. Material properties are not explicit network inputs; information about the layered-medium response is carried by the supplied stress fields. Accordingly, the framework evaluates temporal velocity prediction under reference-stress conditioning, without independently advancing the stress field.

The framework is evaluated through four numerical cases involving one-dimensional SH-wave propagation in four-layer elastic soil columns with fixed total depths of 50 and 280 m. Layer thicknesses and shear-wave velocities are varied to generate heterogeneous profiles, with thickness variations constrained to preserve total depth. The columns are excited by third-order Gaussian-derivative wavelets with selected central frequencies of 0.5, 3.5, and 6 Hz across the four cases, allowing the framework to be evaluated for responses with different characteristic wavelengths and temporal oscillation scales. The temporal extrapolation task is evaluated on test profiles by recursively predicting velocity beyond the training time window while continuing to supply reference stress fields and prescribed excitation, with additional prediction durations up to twice the training duration. Model performance is assessed by comparing with numerical reference solutions, including full-wavefield velocity predictions, surface and base-velocity time histories, temporal error-growth curves, and distributions of the normalized root-mean-square error (RMSE). Together, these comparisons assess the accuracy and temporal generalization of the coupled TI-DeepONet–FNO framework across variable layered profiles, with reference stress fields supplied throughout prediction.

\section{Methodology}
\label{sec:methodology}

\subsection{Governing equations and boundary conditions}
\label{subsec:elastic_wave_equation}
The propagation of stress waves in a solid medium is governed by the balance of linear momentum, expressed in terms of the displacement vector $\boldsymbol{u}$ and Cauchy stress tensor $\boldsymbol{\sigma}$:
\begin{equation}
    \nabla \cdot \boldsymbol{\sigma} + \rho \boldsymbol{f} = \rho \boldsymbol{u_{tt}},
\end{equation}
where $\rho$ is the mass density and $\boldsymbol{f}$ is the body-force vector per unit mass. In the present study, no body force acts within the soil column; the excitation is applied through the base boundary condition described below. For small-strain linear elasticity, the stress tensor is related to the displacement gradient through:
\begin{equation}
    \boldsymbol{\sigma} = \mathsf{C} : \boldsymbol{\varepsilon},
    \;\; \text{where,}\quad
    \boldsymbol{\varepsilon}
    = \frac{1}{2}
    \left[
        \nabla \boldsymbol{u}
        + (\nabla \boldsymbol{u})^{\mathsf{T}}
    \right],
    \label{eq:constitutive_3d}
\end{equation}
and $\mathsf{C}$ is the fourth-order elasticity tensor, and $\boldsymbol{\varepsilon}$ is the infinitesimal strain tensor. For isotropic elasticity, $\mathsf{C}$ is determined by the Lam\'e parameters $\lambda$ and $\mu$, which may vary between layers.

For vertically propagating, horizontally polarized shear (SH) waves in horizontally layered elastic media, the governing equations reduce to a one-dimensional first-order velocity--stress system. Let $z$ denote depth measured positively downward from the ground surface, with $z=0$ at the surface and $z=H$ at the base. The horizontal displacement is denoted by $u(z,t)$, and the corresponding velocity and shear stress are defined as:
\begin{equation}
    v(z,t) = \frac{\partial u(z,t)}{\partial t},
    \qquad
    \sigma(z,t) = G(z)\frac{\partial u(z,t)}{\partial z}.
    \label{eq:velocity_stress_definitions}
\end{equation}
In the absence of interior body forces, the governing equations are:
\begin{align}
    \rho(z)\frac{\partial v}{\partial t}
    &= \frac{\partial \sigma}{\partial z},
    \label{eq:momentum} \\
    \frac{\partial \sigma}{\partial t}
    &= G(z)\frac{\partial v}{\partial z},
    \label{eq:constitutive}
\end{align}
where $\rho(z)$ and $G(z)$ denote the mass density and shear modulus, respectively, with shear-wave velocity $c_s(z)=\sqrt{G(z)/\rho(z)}$.
The velocity--stress system is supplemented by zero initial velocity and stress throughout the soil column,
\begin{equation}
    v(z,0)=0,
    \qquad
    \sigma(z,0)=0,
    \qquad 0 \leq z \leq H.
    \label{eq:initial_conditions}
\end{equation}
A traction-free condition is imposed at the ground surface,
\begin{equation}
    \sigma(0,t)=0.
    \label{eq:free_surface}
\end{equation}
At the base, the prescribed incident-wave velocity history $v_{\mathrm{inc}}(t)$ is introduced through a traction input combined with an absorbing boundary condition~\cite{lysmer1969finite}:
\begin{equation}
    \sigma(H,t)=q(t)-\rho_b c_{s,b}\,v(H,t),
    \qquad
    q(t)=2\rho_b c_{s,b}\,v_{\mathrm{inc}}(t),
    \label{eq:base_boundary}
\end{equation}
where $\rho_b$ and $c_{s,b}$ are the density and shear-wave velocity of the base layer. The prescribed excitation traction $q(t)$ introduces the incident wave, while the absorbing contribution $-\rho_b c_{s,b}v(H,t)$ opposes the base velocity and absorbs waves traveling out of the column through the base. Thus, $\sigma(H,t)$ denotes the total shear stress at the base and is distinct from the prescribed excitation $q(t)$. The total base velocity $v(H,t)$ is determined by the solution rather than prescribed. Because the model is strictly one-dimensional, no lateral boundary conditions are required.

\subsection{Proposed framework}
\label{subsec:proposed_framework}
The preceding subsection defines the coupled first-order equations governing the evolution of velocity and stress in the layered soil column. A complete forward solution determines both fields from the initial conditions, material properties, and prescribed boundary excitation. The present framework instead learns to advance the velocity field using reference stress fields supplied throughout prediction. After initialization with reference velocity fields at the first two time steps, subsequent velocity inputs are obtained from the model's own predictions. Reference stress fields and the prescribed base traction $q(t)$ remain available at each step, including beyond the training interval. The learning task is therefore formulated as stress-conditioned temporal prediction of velocity, rather than autonomous evolution of the coupled velocity--stress system.

The framework learns a one-step neural operator that maps velocity and reference stress fields, together with the prescribed base traction and temporal coordinates at the current and preceding time levels, to the next velocity field:
\begin{equation}
    \mathcal{G}_{\theta}:
    \left(
        v^{n}, v^{n-1},
        \sigma_{\mathrm{ref}}^{n}, \sigma_{\mathrm{ref}}^{n-1},
        q^{n}, q^{n-1},
        t_n, t_{n-1}
    \right)
    \mapsto \widehat{v}^{\,n+1},
    \label{eq:operator_mapping}
\end{equation}
where $t_n=n\Delta t$, $\Delta t$ is the time-step size, and $\theta$ denotes the trainable parameters of the operator. The velocity and stress inputs are spatial fields sampled along the depth coordinate $z$, whereas $q^n=q(t_n)$ and $q^{n-1}$ are scalar traction inputs, supplied to the network after min--max normalization. Spatial arguments are suppressed in Eq.~\eqref{eq:operator_mapping} for readability. The subscript ``ref'' identifies stresses obtained from the reference simulation, and the hat denotes a predicted velocity field. During rollout, $v^n$ and $v^{n-1}$ represent the available velocity fields: reference values for initialization and model predictions thereafter. The two-level input history provides temporal information for the learned update. Temporal extrapolation is evaluated by applying this operator beyond the time interval used to construct the training pairs.

\subsubsection{Time-Integrator Embedded Deep Operator Network}
\label{subsubsec:ti_deeponet}
Since the learning task requires sequential velocity predictions, we employ the time-integrator-embedded Deep Operator Network (TI-DeepONet)~\cite{NAYAK2026118960} as the temporal predictor in the coupled framework. The TI-DeepONet architecture augments the standard DeepONet framework~\cite{lu2021learning} by embedding a numerical time-integration structure directly into the network design, drawing inspiration from neural ordinary differential equations (NODEs)~\cite{chen2018neural}. In the canonical DeepONet, an operator $\mathcal{G}: \mathcal{U} \rightarrow \mathcal{V}$ mapping between infinite-dimensional function spaces is approximated via the decomposition:
\begin{equation}
    \mathcal{G}_\theta(u)(\mathbf{y}) \approx \sum_{k=1}^{p} b_k(u;\, \theta_{\mathrm{branch}}) \cdot t_k(\mathbf{y};\, \theta_{\mathrm{trunk}}),
    \label{eq:deeponet}
\end{equation}
where the \emph{branch network} encodes the input function $u$ (evaluated at a fixed set of sensor locations) into coefficients $\{b_k\}_{k=1}^{p}$, and the \emph{trunk network} produces a set of basis functions $\{t_k\}_{k=1}^{p}$ evaluated at the query location $\mathbf{y}$. Here, $u$ denotes a generic input function rather than the displacement field introduced above, and $p$ is the number of terms in the expansion. The output is then reconstructed as a linear combination of these basis functions weighted by the branch outputs.

TI-DeepONet extends this paradigm by explicitly incorporating a numerical time-in\-te\-gra\-tion scheme into the prediction process. Rather than treating time as merely another coordinate in the trunk network, TI-DeepONet embeds a time-stepping scheme that respects the sequential nature of dynamical systems. This architectural choice is intended to (i) support stable autoregressive rollouts, (ii) preserve temporal coherence in the predicted velocity fields, and (iii) integrate a learned velocity-rate function with an established numerical time-stepping scheme.

\paragraph{Multi-branch encoding of the input state}
To accommodate the heterogeneous inputs, we employ four independent branch sub-networks, one each for the reference stress, velocity, prescribed base traction, and temporal-coordinate channels. Each sub-network ingests the concatenation of its input values at the current and preceding time levels; for instance, the stress branch receives $[\sigma_{\mathrm{ref}}^{n},\sigma_{\mathrm{ref}}^{n-1}]$. The velocity branch receives $[v^n,v^{n-1}]$, while the traction and temporal-coordinate branches receive $[q^n,q^{n-1}]$ and $[t_n,t_{n-1}]$, respectively. Denoting the branch encodings by $b_\sigma$, $b_v$, $b_q$, and $b_\tau$, respectively, the individual embeddings are fused through a multiplicative--additive interaction followed by concatenation with the temporal embedding and a learnable linear transformation:
\begin{equation}
    \beta = \Bigl[b_\sigma \odot b_v + b_q,\;b_\tau\Bigr] W_c,
    \label{eq:branch_fusion}
\end{equation}
where $\odot$ denotes the elementwise (Hadamard) product, $[\cdot\,,\cdot]$ denotes concatenation along the latent dimension, and $W_c$ is a learnable linear map that produces the final branch coefficient vector $\beta \in \mathbb{R}^{p}$. The trunk network $\{t_k(\mathbf{y})\}_{k=1}^{p}$ is evaluated on the spatial query coordinates, which in the present one-dimensional setting correspond to depth $z$. The depth coordinate is first lifted through a Fourier feature encoding $\gamma(z)=[\sin(\boldsymbol{\omega}z),\cos(\boldsymbol{\omega}z)]$ over a bank of fixed frequencies $\boldsymbol{\omega}$, and processed using an adaptive wavelet activation of the form $w_1\sin(\omega a)+w_2\cos(\omega a)$ with trainable $w_1$, $w_2$, and $\omega$~\cite{zhao2023pinnsformer}. Here, $a$ denotes the scalar input to the activation function; its trainable frequency parameter is distinct from the fixed frequencies used in the Fourier encoding. This spectral parameterization of the trunk is intended to facilitate the representation of the oscillatory structure of the wavefield. 

\paragraph{Embedded time integration}
The key departure from the canonical DeepONet is that the network does not directly emit the next-step velocity field; instead, it parameterizes a velocity-rate function used within the time-integration scheme:
\begin{equation}
    r_\theta^n(z)=\mathcal{N}_\theta
    \left(v^n,v^{n-1},\sigma_{\mathrm{ref}}^n,\sigma_{\mathrm{ref}}^{n-1},
        q^n,q^{n-1},t_n,t_{n-1}\right)(z).
    \label{eq:ti_rhs}
\end{equation}
The learned rate is used to construct a preliminary next-step velocity through an embedded classical fourth-order Runge--Kutta (RK4) scheme. Writing $\mathcal{N}_\theta(\cdot)$ compactly as a function of the velocity state, the RK4 stage evaluations are:
\begin{align}
    k_1 &= \mathcal{N}_\theta(v^n), \notag\\
    k_2 &= \mathcal{N}_\theta
    \left(v^n+\tfrac12\Delta t\,k_1\right), \notag\\
    k_3 &= \mathcal{N}_\theta
    \left(v^n+\tfrac12\Delta t\,k_2\right),
    \label{eq:rk4_stages}\\
    k_4 &= \mathcal{N}_\theta
    \left(v^n+\Delta t\,k_3\right), \notag
\end{align}
where the reference stress and prescribed traction inputs are held fixed across stages and the temporal coordinate is offset consistently by $\tfrac12\Delta t$ and $\Delta t$ at the intermediate and final stages. The additional inputs and stage-time dependence are suppressed in this compact notation. The preliminary velocity prediction at the next time level is obtained by the classical RK4 update,
\begin{equation}
    \widetilde{v}^{\,n+1}=v^n+\frac{\Delta t}{6}\left[k_1+2k_2+2k_3+k_4\right],
    \label{eq:rk4_update}
\end{equation}
with $\Delta t$ the discrete time step. By learning a velocity-rate function and integrating it through the RK4 update, the temporal predictor combines a learned representation of the dynamics with an established numerical time-stepping structure. Its performance within the coupled predictor--corrector framework is evaluated through the rollout results presented in Section~\ref{sec:results}.

\subsubsection{Fourier Neural Operator}
\label{subsubsec:fno}
While TI-DeepONet provides the temporal prediction stage, the FNO supplies a spatial correction to the predicted velocity field. To this end, we incorporate the Fourier Neural Operator (FNO)~\cite{li2020fourier}, which has been developed to learn solution operators for partial differential equations. The FNO parameterizes an integral kernel in Fourier space, enabling global spatial interactions that represent long-range dependencies. The core building block of the FNO is the Fourier layer, defined as:
\begin{equation}
    h^{(l+1)}(z)=\phi\left(W^{(l)}h^{(l)}(z)+\mathcal{F}^{-1}\left[R^{(l)} \cdot\mathcal{F}\left(h^{(l)}\right)\right] (z)\right),
    \label{eq:fourier_layer}
\end{equation}
where $h^{(l)}$ denotes the hidden representation at layer $l$, $\mathcal{F}$ and $\mathcal{F}^{-1}$ denote the Fourier transform and its inverse, respectively, $R^{(l)}$ is a learnable weight tensor applied in the spectral domain (truncated to the lowest $k_{\max}$ Fourier modes), $W^{(l)}$ is a local linear transformation, and $\phi$ is a pointwise nonlinear activation function. The spectral convolution via $R^{(l)}$ enables the FNO to efficiently capture global spatial patterns, while the local path through $W^{(l)}$ acts on the hidden features without Fourier-mode truncation.

For wave propagation problems, this global receptive field allows the correction at a given depth to depend on features throughout the soil column. This provides a mechanism for learning spatial corrections associated with wavefronts, reflections, and interference patterns.

\subsubsection{TI-DeepONet Coupled with FNO}
\label{subsubsec:coupled_architecture}
To combine the complementary strengths of both architectures, we propose a coupled TI-DeepONet--FNO framework that combines the temporal integration capabilities of TI-DeepONet with the spatial correction capabilities of the FNO. The coupling strategy is designed to address the distinct challenges of stress-conditioned temporal extrapolation and spatial pattern learning within a unified, end-to-end trainable architecture.

\begin{figure}[htbp!]
    \centering
    \includegraphics[width=\linewidth]{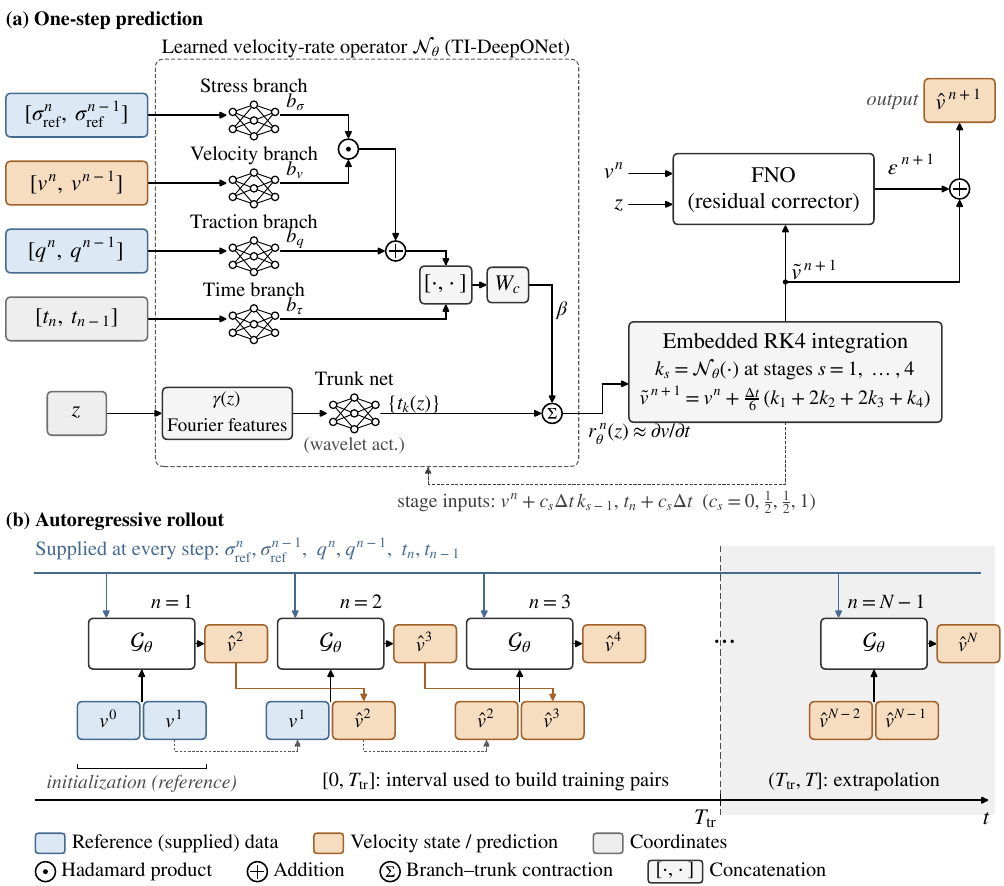}
    \caption{Coupled TI-DeepONet--FNO framework. (a) One-step prediction: four branch networks encode the two-time-level inputs, which are fused as $\beta=[\,b_\sigma\odot b_v+b_q,\;b_\tau\,]W_c$ (Eq.~\eqref{eq:branch_fusion}) and contracted with the trunk basis $\{t_k(z)\}$ to give the learned velocity rate $r_\theta^{\,n}(z)$ (Eq.~\eqref{eq:ti_rhs}). The operator $\mathcal{N}_\theta$ is evaluated at the four stages of the embedded RK4 scheme (Eqs.~\eqref{eq:rk4_stages}--\eqref{eq:rk4_update}) to obtain the preliminary velocity $\tilde v^{\,n+1}$, which the FNO corrects with the residual $\epsilon^{\,n+1}$ (Eqs.~\eqref{eq:fno_correction}--\eqref{eq:refined_prediction}). (b) Autoregressive rollout: the first two velocity fields are taken from the reference solution; thereafter, the corrected predictions $\hat v$ are fed back into the two-level velocity input, while reference stresses, base traction, and time coordinates are supplied at every step, including beyond $T_{\mathrm{tr}}$.}
    \label{fig:TI-DON-FNO_schematic}
\end{figure}

\paragraph{Architecture overview}
The proposed hybrid architecture operates through a predictor--corrector paradigm, as illustrated in Figure~\ref{fig:TI-DON-FNO_schematic}. In the \emph{prediction stage}, the TI-DeepONet advances the velocity field to a preliminary estimate $\tilde{v}^{\,n+1}$, conditioned on the supplied reference stress and prescribed traction inputs, at the next time level through the embedded RK4 integrator of Equations~\eqref{eq:rk4_stages}--\eqref{eq:rk4_update}. This preliminary prediction is subsequently refined by the FNO correction stage.

In the subsequent \emph{correction stage}, the FNO acts as a residual refinement module. The FNO takes as input a three-channel field comprising the current velocity $v^n$, the preliminary prediction $\tilde{v}^{\,n+1}$, and the spatial coordinate grid $z$, and learns to predict a correction term $\epsilon^{n+1}$:
\begin{equation}
    \epsilon^{n+1}(z)=\mathcal{G}_{\theta_{\mathrm{FNO}}}\left(v^n,\tilde{v}^{\,n+1},z\right)(z).
    \label{eq:fno_correction}
\end{equation}
The explicit inclusion of the coordinate channel $z$ supplies the FNO with a positional reference, enabling spatially localized corrections. The refined velocity field is then obtained by adding this learned correction to the preliminary prediction:
\begin{equation}
    \hat{v}^{\,n+1}(z)=\tilde{v}^{\,n+1}(z)+\epsilon^{n+1}(z),
    \label{eq:refined_prediction}
\end{equation}
where $\hat{v}^{\,n+1}$ denotes the final refined prediction. This residual learning strategy allows the FNO to focus specifically on correcting the errors in the TI-DeepONet prediction rather than learning the full solution mapping from scratch. The corrected velocity is used as an input to subsequent prediction steps.

\paragraph{Rationale for the predictor-corrector design}
The inclusion of both $v^n$ and $\tilde{v}^{\,n+1}$ as inputs to the FNO serves a specific purpose. By providing access to the current velocity field $v^{n}$, the FNO can leverage information about the existing wavefield structure, including wavefront locations, amplitude distributions, and spatial gradients, to inform its correction. Meanwhile, the preliminary prediction $\tilde{v}^{n+1}$ provides an estimate of the next-step velocity, enabling the FNO to learn a \emph{state-dependent} correction that accounts for how prediction errors manifest differently across various wavefield configurations. This design draws inspiration from classical predictor--corrector time integration schemes, in which a prediction is followed by a correction. Here, the correction is learned and applied once at each time step. Although multilayer perceptrons can exhibit spectral bias~\cite{rahaman2019spectral}, the present TI-DeepONet trunk already incorporates Fourier features and oscillatory activations. The FNO therefore provides an additional mechanism for spatial refinement, without assuming that the remaining prediction errors arise solely from spectral bias or are confined to high-frequency content.

\paragraph{End-to-end training}
The entire coupled architecture is trained jointly in an end-to-end manner. During training, gradients are backpropagated through both the FNO correction stage and the RK4-embedded TI-DeepONet prediction stage, allowing the two components to co-adapt. To promote robustness to errors in the velocity inputs during autoregressive rollout, we perturb the input velocity field $v^{n}$ with zero-mean Gaussian noise, $v^{n} \leftarrow v^{n} + \eta$, $\eta \sim \mathcal{N}(0, s^2 I)$, whose scale $s$ is annealed over the course of training. The training pairs are assembled by sweeping the two-time-level stencil across all admissible time levels of the training trajectories, restricting both the input and target times to $[0,T_{\mathrm{tr}})$. The resulting pairs are pooled, and the mean squared error is minimized over all pairs and sampled depth locations, evaluated on the min--max normalized fields used for training:
\begin{equation}
\mathcal{L}(\theta)=\frac{1}{N_p N_z}\sum_{i=1}^{N_p}\sum_{j=1}^{N_z}\left(\hat{v}^{\,n+1}_{(i)}(z_j)-v^{\,n+1}_{\mathrm{ref},(i)}(z_j)\right)^{2},
\label{eq:loss_function}
\end{equation}
where $N_p$ denotes the total number of input--output pairs, each associated with a sample trajectory and a time level, and $N_z$ is the number of sampled depth locations. For each pair $i$, $n$ denotes its current time index, $\hat{v}_{(i)}^{n+1}$ is the final corrected velocity prediction, and $v_{\mathrm{ref},(i)}^{n+1}$ is the corresponding reference velocity field obtained from the numerical simulation. The training loss is evaluated on the min--max normalized velocity fields. Prediction errors defined in Section~\ref{subsec:error_metrics} are evaluated after restoring the predicted and reference velocities to physical units. The joint training strategy optimizes the final corrected velocity rather than the intermediate velocity-rate estimate or preliminary prediction separately, allowing the predictor and corrector to adapt to one another.

\paragraph{Summary of the coupled operator}
The complete coupled operator mapping can be expressed as:
\begin{equation}
    \mathcal{G}_{\theta}^{\mathrm{coupled}}:\left(v^n,v^{n-1},\sigma_{\mathrm{ref}}^n,\sigma_{\mathrm{ref}}^{n-1},q^n,q^{n-1},t_n,t_{n-1}\right)\mapsto \hat{v}^{\,n+1},
    \label{eq:coupled_operator}
\end{equation}
with the full trainable parameter set $\theta = \{\theta_{\mathrm{TI\text{-}DON}}, \theta_{\mathrm{FNO}}\}$. This hybrid framework combines the embedded time integration of TI-DeepONet with the residual spatial correction of the FNO to predict the evolution of the velocity field in layered elastic media. During rollout, the corrected velocity predictions are fed back into the two-level input history, while reference stress fields and prescribed traction inputs continue to be supplied.

\subsection{Error metrics}
\label{subsec:error_metrics}
Prediction accuracy is evaluated using peak-normalized root-mean-square error (NRMSE) on the predicted and reference velocity fields after restoring them to physical units. Let $z_i$, with $i=1,\ldots,N_z$, denote the sampled depth locations, and let $t_n$, with $n=1,\ldots,N_t$, denote the sampled time levels within the reported evaluation window $[0,T]$. For each test profile $k$, the normalization amplitude is the maximum absolute reference velocity over this depth--time window:
\begin{equation}
A^{(k)} = \max_{\substack{1\leq i\leq N_z\\1\leq n\leq N_t}} \left|v_{\mathrm{ref}}^{(k)}(z_i,t_n)\right|,
\label{eq:peak_amp}
\end{equation}
where $v_{\mathrm{ref}}^{(k)}$ denotes the reference velocity field for test profile $k$. The amplitude $A^{(k)}$ is positive for all evaluated profiles and is used consistently to normalize the following metrics.

The NRMSE over the full evaluation window is defined as:
\begin{equation}
\varepsilon_{\mathrm{NRMSE}}^{(k)} = \frac{1}{A^{(k)}}\left[\frac{1}{N_tN_z}\sum_{n=1}^{N_t}\sum_{i=1}^{N_z}\left(\hat{v}^{(k)}(z_i,t_n) - v_{\mathrm{ref}}^{(k)}(z_i,t_n) \right)^2 \right]^{1/2},
\label{eq:rel_l2_field}
\end{equation}
where $\hat{v}^{(k)}$ denotes the predicted velocity field. The first two velocity fields supplied for initialization are included in the full-window calculation. Their residuals are zero, and their time levels remain included in the averaging. To evaluate prediction beyond the training interval, define $\mathcal{I}_{\mathrm{ex}} =\{n:T_{\mathrm{tr}}\leq t_n\leq T\}$ and $N_{\mathrm{ex}}=|\mathcal{I}_{\mathrm{ex}}|$, where $T_{\mathrm{tr}}$ denotes the first time level outside the training interval. The extrapolation-only NRMSE is:
\begin{equation}
\varepsilon_{\mathrm{NRMSE,ex}}^{(k)} = \frac{1}{A^{(k)}} \left[ \frac{1}{N_{\mathrm{ex}}N_z} \sum_{n\in\mathcal{I}_{\mathrm{ex}}}\sum_{i=1}^{N_z} \left(\hat{v}^{(k)}(z_i,t_n) - v_{\mathrm{ref}}^{(k)}(z_i,t_n) \right)^2 \right]^{1/2}.
\label{eq:rel_l2_extrap}
\end{equation}
Only the RMS discrepancy is restricted to the extrapolation interval; the normalizer remains the reference peak over the full evaluation window. Both metrics therefore express errors relative to the same event amplitude. Each test profile yields one full-window NRMSE and one extrapolation-only NRMSE. The reported means and medians are calculated across these profile-wise values within each test case.

For an individual station $z_\star\in\{0,H\}$, corresponding to the surface or base, the station-history NRMSE is:
\begin{equation}
\varepsilon_{\mathrm{NRMSE}}^{(k)}(z_\star) = \frac{1}{A^{(k)}} \left[ \frac{1}{N_t} \sum_{n=1}^{N_t} \left( \hat{v}^{(k)}(z_\star,t_n) - v_{\mathrm{ref}}^{(k)}(z_\star,t_n) \right)^2 \right]^{1/2}.
\label{eq:rel_l2_station}
\end{equation}

For the pointwise error panels accompanying the surface and base velocity histories, the absolute prediction discrepancy is normalized by the peak absolute reference surface velocity over the evaluation window:
\begin{equation}
p^{(k)}(z_\star,t_n) = 100\, \frac{ \left|\hat{v}^{(k)}(z_\star,t_n) -v_{\mathrm{ref}}^{(k)}(z_\star,t_n)\right|}{\displaystyle\max_{1\leq m\leq N_t}\left|v_{\mathrm{ref}}^{(k)}(0,t_m)\right|},
\qquad z_\star\in\{0,H\}.
\label{eq:pointwise_error}
\end{equation}
These pointwise errors are expressed as percentages and use the same surface reference peak for both stations. Unlike the station-history NRMSE, they involve no averaging over time.

The temporal variation in prediction error is characterized by the instantaneous NRMSE, evaluated over depth at each time level:
\begin{equation}
e^{(k)}(t_n) = \frac{1}{A^{(k)}} \left[ \frac{1}{N_z} \sum_{i=1}^{N_z} \left(\hat{v}^{(k)}(z_i,t_n) - v_{\mathrm{ref}}^{(k)}(z_i,t_n)\right)^2\right]^{1/2}.
\label{eq:inst_error}
\end{equation}
The instantaneous NRMSE measures the spatial RMS discrepancy at each time level relative to the fixed reference amplitude $A^{(k)}$; it is not accumulated over time. The curves reported in Section~\ref{sec:results} show the across-profile mean, $\mu(t_n)=\dfrac{1}{N_s}\sum_{k=1}^{N_s}e^{(k)}(t_n)$, where $N_s$ is the number of test profiles. The shaded band extends one population standard deviation above and below the mean, with its lower bound clipped at zero. The standard deviation is calculated across profiles using division by $N_s$. All NRMSE values reported in the text, tables, and figures are expressed as percentages.

\section{Problem Setup}
\label{sec:problem_statement}
\begin{figure}[!ht]
    \centering
    \includegraphics[width=0.95\linewidth]{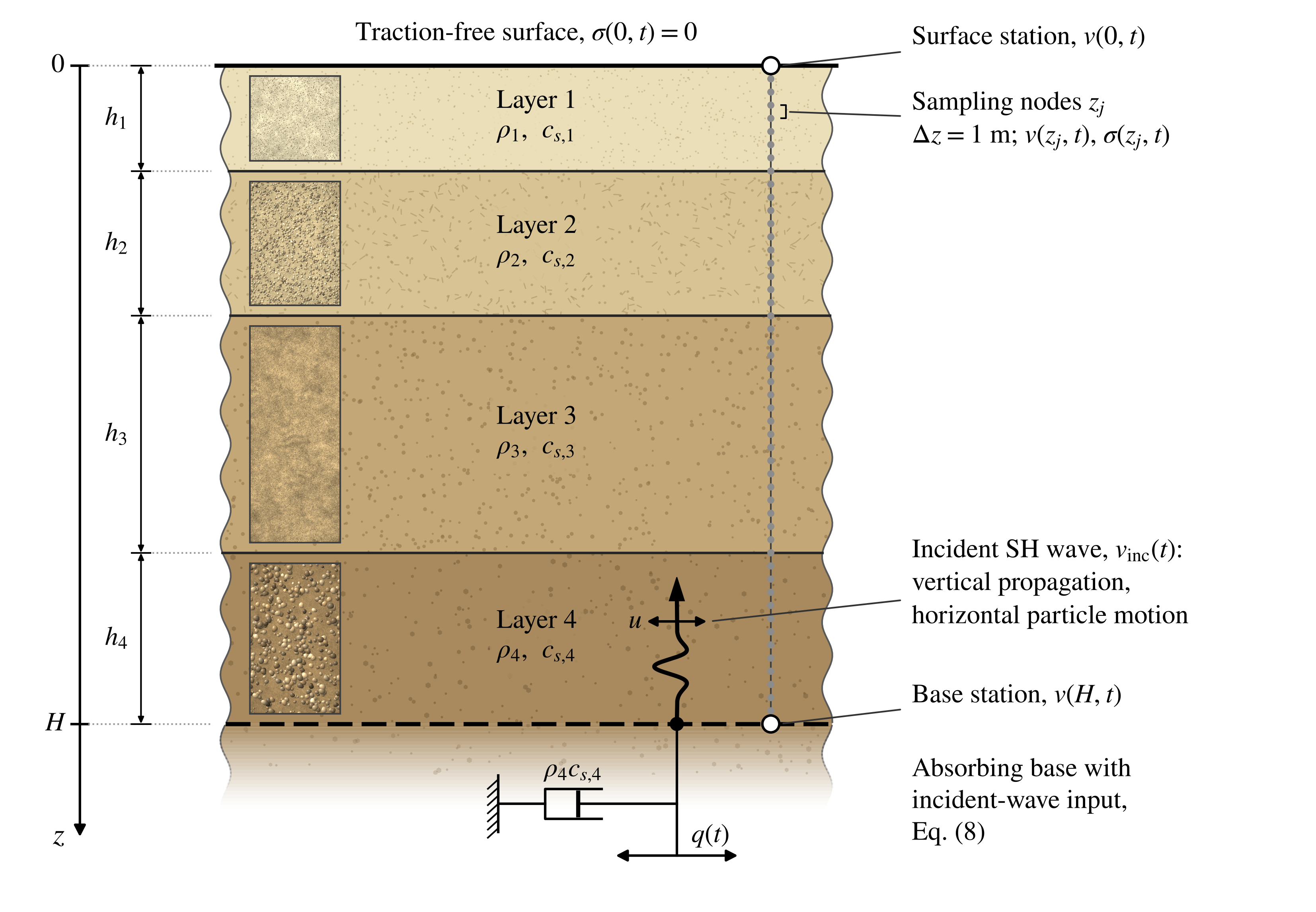}
    \caption{General configuration of the four-layer elastic soil column, with layer thicknesses $h_i$, densities $\rho_i$, shear-wave velocities $c_{s,i}$ ($c_{s,i}<c_{s,i+1}$), and total depth $H=\sum_{i=1}^{4}h_i$. Depth $z$ increases downward from the traction-free surface. Layer~4 extends semi-infinitely and is truncated at $z=H$, where the absorbing boundary (dashpot $\rho_4 c_{s,4}$) and the incident-wave traction $q(t)$ of Eq.~\eqref{eq:base_boundary} are applied, with $\rho_b=\rho_4$ and $c_{s,b}=c_{s,4}$. Fields are sampled at $\Delta z=1$ or $0.25$~m; open circles denote the surface and base stations. Particle motion is drawn in-plane for clarity. The schematic applies to both configurations in Tables~\ref{tab:shallow_layer} and~\ref{tab:deep_layer}; it is not to scale, and the soil swatches are illustrative.}
    \label{fig:problem_configuration}
\end{figure}

To evaluate the performance of the proposed TI-DeepONet--FNO framework for stress-conditioned velocity prediction, we conduct one-dimensional computational experiments involving vertically propagating SH waves in horizontally layered elastic soil columns. Two fixed total depths, $H=50$~m and $H=280$~m, are considered. Each model consists of four homogeneous layers, where layer $i$ is characterized by its thickness $h_i$ and shear-wave velocity $c_{s,i}$. The general configuration is illustrated in Figure~\ref{fig:problem_configuration}, and the baseline layer properties are listed in Tables~\ref{tab:shallow_layer} and~\ref{tab:deep_layer}. The governing equations and boundary conditions follow the formulation in Section~\ref{subsec:elastic_wave_equation}.
A traction-free condition is imposed at the ground surface, while the base boundary combines incident-wave excitation with absorption of outgoing waves, as described in Section~\ref{subsec:elastic_wave_equation}. To assess the framework's performance under different excitation frequencies, the models are excited using third-order Gaussian-derivative (DOG3) wavelets prescribed as incident velocity histories. The deep-column cases use central frequencies of 0.5 and 3.5~Hz, whereas the shallow-column cases use 3.5 and 6~Hz. 

Furthermore, to represent variability in the layered profiles, the shear-wave velocities $c_{s,i}$ and layer thicknesses $h_i$ are varied using a constrained discrete sampling approach. For each layer, candidate values are generated within a $\pm20\%$ range of the baseline properties, on discrete grids with increments $\Delta c_s=5$~m/s and $\Delta h=1$~m. This approach defines a discrete candidate set for each property of layer $i$:
\begin{equation}
    \mathcal{P}_i=\left\{k\Delta p:k\in\mathbb{Z},\quad\left\lceil\frac{(1-\delta)p_{i,0}}{\Delta p}\right\rceil\leq k \leq\left\lfloor\frac{(1+\delta)p_{i,0}}{\Delta p}\right\rfloor\right\},
    \label{eq:property_sampling}
\end{equation}
where $p_i$ denotes either $c_{s,i}$ or $h_i$, $p_{i,0}$ is its baseline value, $\delta=0.20$ is the relative perturbation bound, and $\Delta p$ is the corresponding increment. The ceiling and floor operators keep the candidate values within the prescribed bounds while maintaining the specified increments. The sampled profiles are restricted to increasing shear-wave velocity with depth, so a strictly increasing velocity constraint ($c_{s,i}<c_{s,i+1}$) is enforced during the assembly of the models. The layer thicknesses additionally satisfy $\sum_{i=1}^{4}h_i=H$, preserving the specified total column depth. From the resulting valid configurations, a subset of $N$ unique models is randomly sampled.
\begin{table}[htbp!]
    \centering
    \caption{Baseline layer properties and discrete candidate ranges
    for the shallow soil column ($H=50$~m; $\delta=0.20$).}
    \label{tab:shallow_layer}
    \begin{tabular}{@{} c c c c c @{}} 
        \toprule
        Layer &
        \shortstack{Baseline\\$h_{i,0}$ [m]} &
        \shortstack{Candidate $h_i$\\range [m]} &
        \shortstack{Baseline\\$c_{s,i,0}$ [m/s]} &
        \shortstack{Candidate $c_{s,i}$\\range [m/s]} \\
        \midrule
        1 & 10 & [8, 12]   & 218 & [175, 260] \\ 
        2 & 10 & [8, 12]   & 301 & [245, 360] \\ 
        3 & 21 & [17, 25]  & 396 & [320, 475] \\ 
        4 & 9  & [8, 10]   & 462 & [370, 550] \\
        \bottomrule
    \end{tabular}  
\end{table}
\begin{table}[htbp!]
    \centering
    \caption{Baseline layer properties and discrete candidate ranges for the deep soil column ($H=280$~m; $\delta=0.20$).}
    \label{tab:deep_layer}
    \begin{tabular}{@{} c c c c c @{}} 
        \toprule
        Layer &
        \shortstack{Baseline\\$h_{i,0}$ [m]} &
        \shortstack{Candidate $h_i$\\range [m]} &
        \shortstack{Baseline\\$c_{s,i,0}$ [m/s]} &
        \shortstack{Candidate $c_{s,i}$\\range [m/s]} \\
        \midrule
        1 & 26  & [21, 31]  & 287  & [230, 340] \\ 
        2 & 66  & [53, 79]  & 480  & [385, 575] \\ 
        3 & 107 & [86, 128] & 719  & [580, 860] \\ 
        4 & 81  & [65, 97]  & 1004 & [805, 1200] \\
        \bottomrule
    \end{tabular}

\medskip
\begin{minipage}{\linewidth}
    \footnotesize
    \textit{Note for Tables~\ref{tab:shallow_layer}
    and~\ref{tab:deep_layer}:}
    Layers are numbered from the ground surface downward. Ranges denote the endpoints of discrete candidate sets with increments $\Delta h=1$~m and $\Delta c_s=5$~m/s. Accepted profiles satisfy $\sum_{i=1}^{4}h_i=H$, where $H=50$~m for the shallow column and $H=280$~m for the deep column, and $c_{s,i}<c_{s,i+1}$ for $i=1,2,3$.
\end{minipage}    
\end{table}

To construct the synthetic dataset, each sampled soil column is excited at its base using a DOG3 incident velocity wavelet through the boundary condition defined in Section~\ref{subsec:elastic_wave_equation}, and the resulting velocity and stress fields are sampled throughout the domain at 1 m and 0.25 m spatial intervals for deep and shallow profiles, respectively. The primary objective is to evaluate the ability of the proposed framework to predict velocity fields beyond the training time window under reference-stress conditioning. Specifically, training pairs are constructed from the initial temporal interval, while test-profile predictions are initialized with the first two reference velocity fields and advanced recursively over the remaining record. Reference stress fields and the prescribed excitation remain inputs throughout prediction. The training interval is $[0,T_{\mathrm{tr}})$, and the extrapolation interval is $[T_{\mathrm{tr}},T]$, with $T_{\mathrm{tr}}$ denoting the first time level outside the training interval. Performance is evaluated using full-wavefield velocity predictions and surface and base velocity histories.

% justify the training interval
Candidate training windows were specified by timestep count. For the shallow 3.5-Hz configuration, preliminary comparisons supported the reported 1-s training window, as discussed in Section~\ref{subsec:shallow_profile_3.5Hz}. The ratio of the characteristic wavelength, $\lambda_i=c_{s,i}/f_c$, to the layer thickness, $h_i$, helps interpret the wave arrivals within the selected windows. When $\lambda_i$ is substantially larger than $h_i$, the layer is thin relative to the characteristic wavelength, and contributions from successive interface interactions may overlap in the time histories. For example, in the shallow 3.5-Hz case, the wavelength in the first layer is much larger than its thickness. The selected training interval includes the dominant arrivals, so the subsequent extrapolation interval primarily tests prediction of the lower-amplitude response. At higher excitation frequencies, the wavelengths become more comparable to the layer thicknesses, and the shorter-duration DOG3 pulses can make individual arrivals more distinguishable. In the shallow 6-Hz case, the selected shorter training interval leaves part of the reflected-wave sequence within the extrapolation interval. These wavelength considerations provide physical context rather than a quantitative criterion for determining the minimum training duration.

\section{Results and Discussion}
\label{sec:results}
The TI-DeepONet–FNO framework is evaluated using four combinations of soil-column depth and excitation frequency to examine temporal extrapolation under different wave-propagation conditions. Surface and base velocity histories assess the reproduction of individual waveforms, while depth–time seismograms evaluate the spatial consistency of the predicted response. Sample-wise normalized RMSEs characterize performance across test profiles, and time-dependent errors describe how prediction discrepancies evolve during rollout. Together, these measures distinguish agreement in the dominant response from errors in weaker arrivals and late-time motion. Particular attention is given to the response beyond the training interval, where velocity predictions are evaluated at times excluded from training, while reference stress fields and the prescribed excitation remain inputs throughout rollout. 
\textcolor{black}{The deep 0.5-Hz, deep 3.5-Hz, and shallow 3.5-Hz cases each contain 500 profiles, and the shallow 6-Hz case contains 1000. Training pairs are drawn from all profiles over $[0, T_{tr})$, with a random 20\% of these pairs held out to monitor training, and all profiles are included in the reported statistics. Because the response over $[T_{tr}, T]$ is excluded from training for every profile, the errors over the extrapolation interval measure prediction at times not represented in training.} 
% \djn{Check usage.}
% The deep 0.5-Hz, deep 3.5-Hz, shallow 3.5-Hz, and shallow 6-Hz cases each contain 100 test profiles. All test profiles are included in the reported statistics.
\begin{table}[htbp!]
\centering
\caption{Summary of peak-normalized RMSEs across the four test cases. Mean and median values are calculated across individual test profiles and reported as percentages for the full evaluation window and the extrapolation interval. Both metrics are normalized by the peak absolute reference velocity of each profile over the full evaluation window. The selected profiles correspond to the lowest, near-median, and highest NRMSEs when ranked over the full evaluation window.}
\label{tab:results_summary}
\resizebox{\textwidth}{!}{%
\begin{tabular}{lcccccccccc}
\toprule
& & & &
\multicolumn{4}{c}{Test-set statistics [\%]} &
\multicolumn{3}{c}{Selected-profile NRMSEs [\%]} \\
\cmidrule(lr){5-8}\cmidrule(lr){9-11}
Profile & $f_c$ [Hz] & Training [s] & Extrapolation [s] &
\multicolumn{2}{c}{Full window} &
\multicolumn{2}{c}{Extrapolation only} &
Lowest & Near-median & Highest \\
\cmidrule(lr){5-6}\cmidrule(lr){7-8}
& & & & Mean & Median & Mean & Median & & & \\
\midrule
Deep    & 0.5 & $[0,4)$   & $[4,8]$
        & 1.88 & 1.85 & 2.59 & 2.54 & 1.4889 & 1.8468 & 2.7121 \\
Deep    & 3.5 & $[0,2)$   & $[2,4]$
        & 0.78 & 0.31 & 0.73 & 0.14 & 0.1997 & 0.3075 & 13.1141 \\
Shallow & 3.5 & $[0,1)$   & $[1,2]$
        & 0.53 & 0.52 & 0.61 & 0.60 & 0.3632 & 0.5244 & 0.9731 \\
Shallow & 6.0 & $[0,0.5)$ & $[0.5,1.5]$
        & 2.31 & 2.31 & 2.82 & 2.82 & 1.9839 & 2.3090 & 2.7020 \\
\bottomrule
\end{tabular}%
}
\end{table}

\subsection{Deep Profile with 0.5 Hz}
\label{subsec:deep_profile_0.5Hz}
\begin{figure}[htbp!]
    \centering
    \includegraphics[width=\linewidth]{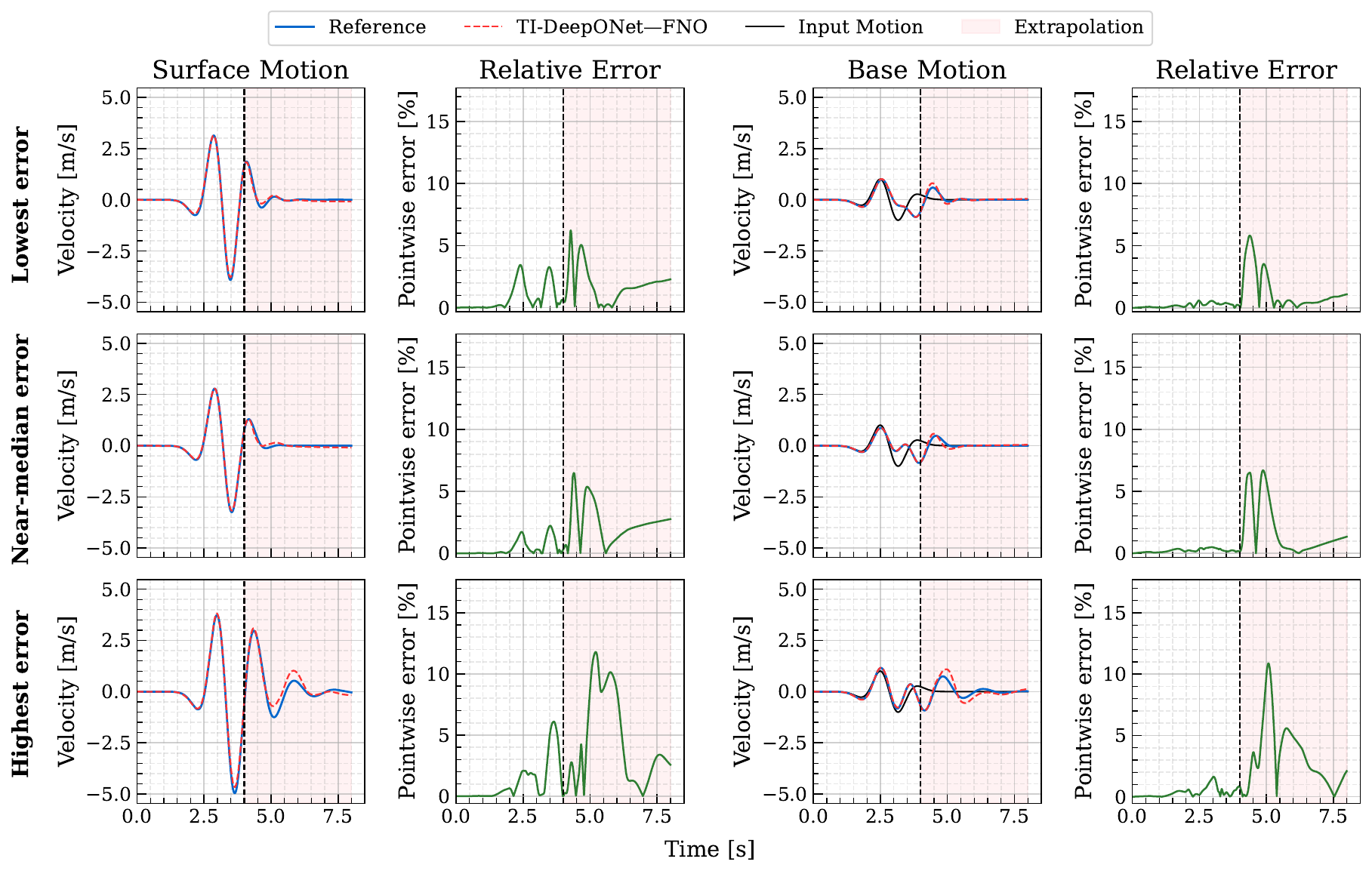}
    \caption{Deep profile with 0.5-Hz excitation: reference and TI-DeepONet--FNO surface and base velocity histories for profiles selected by their lowest, near-median, and highest NRMSEs over the full evaluation window. Pointwise errors at both stations are expressed as percentages of each profile's peak absolute reference surface velocity over the displayed evaluation window. The training interval is $[0,4)$~s, and shading marks the extrapolation interval $[4,8]$~s. }
    \label{fig:deep_05Hz_surface}
\end{figure}

\begin{figure}[htbp!]
    \centering
    \includegraphics[width=\linewidth]{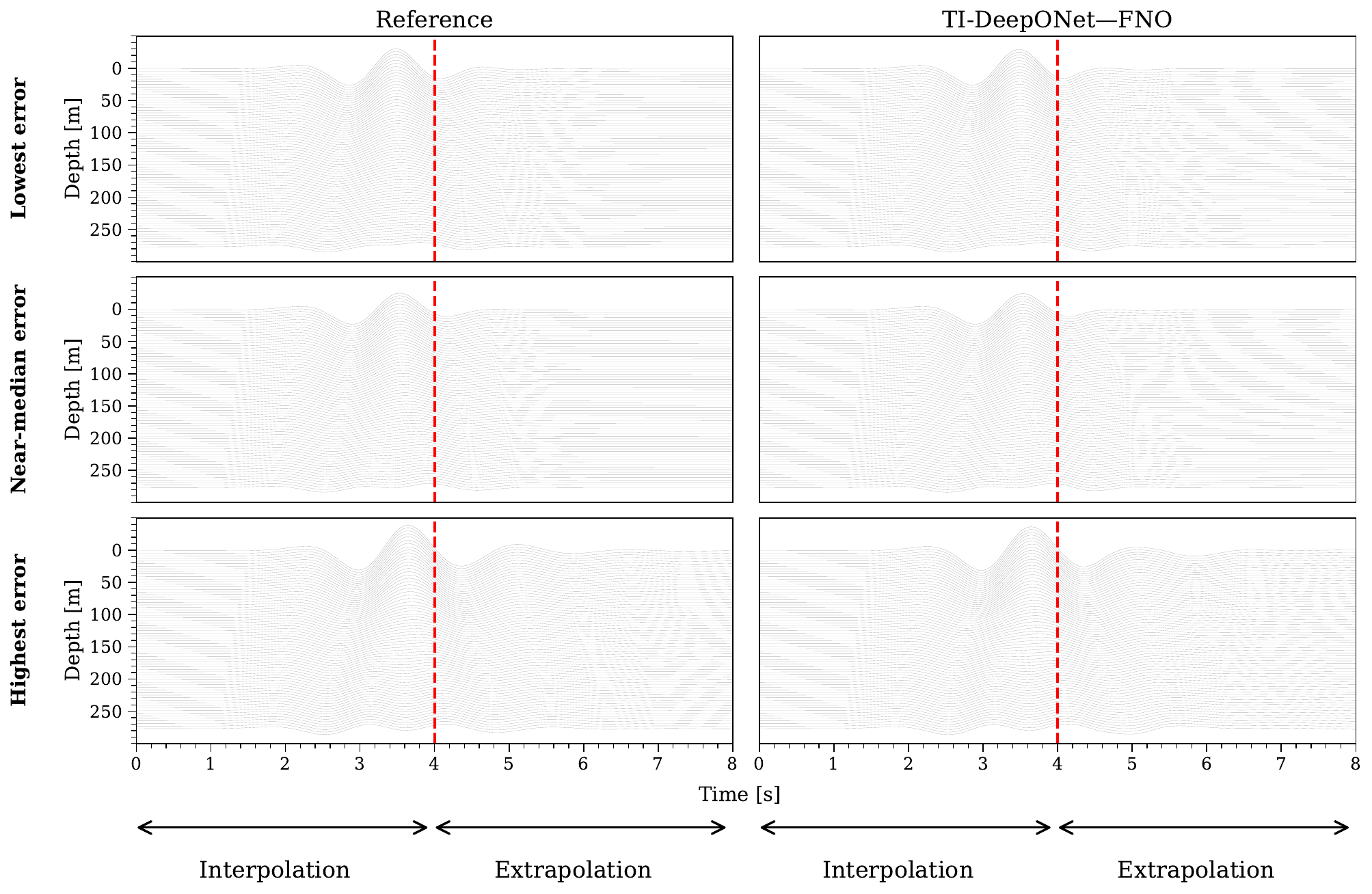}
    \caption{Deep profile with 0.5-Hz excitation: reference and TI-DeepONet--FNO depth--time velocity seismograms for the same profiles shown in Figure~\ref{fig:deep_05Hz_surface}. Dashed vertical lines mark the first extrapolation time level at $t=4$~s; the displayed evaluation window extends to $t=8$~s. Velocities are shown in physical units.}
    \label{fig:deep_05Hz_seismogram}
\end{figure}

The deep-profile case with 0.5-Hz excitation provides a baseline for evaluating temporal extrapolation of a comparatively smooth, long-period response. Figures~\ref{fig:deep_05Hz_surface} and~\ref{fig:deep_05Hz_seismogram} show that TI-DeepONet--FNO reproduces the principal features of the surface and base waveforms and their depth-dependent evolution for the selected profiles. The agreement extends beyond the training--extrapolation boundary at $t=4$~s, including the later response associated with returning wave energy. This is significant because successful velocity prediction requires the model to reproduce the response throughout the column, rather than reproduce only the principal surface pulse. The displayed predictions show no obvious waveform discontinuity at the transition to extrapolation, supporting continuity of the predicted response over the evaluated interval.

\begin{figure}[htbp!]
    \centering
    \includegraphics[width=\linewidth]{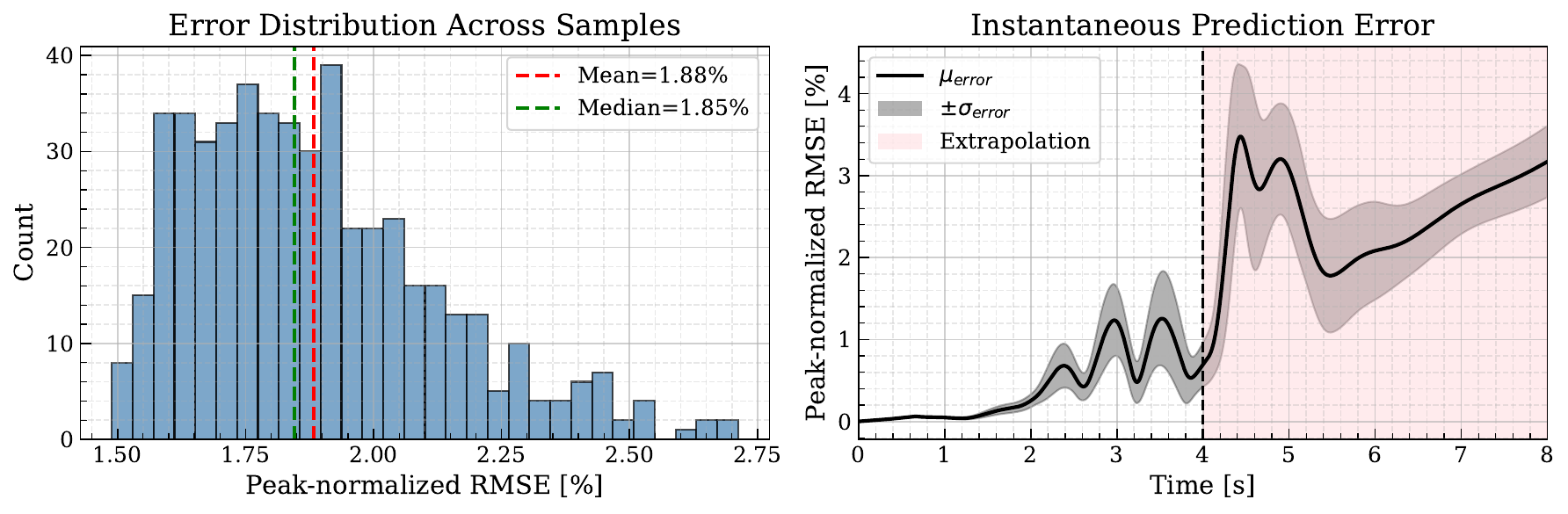}
    \caption{Deep profile with 0.5-Hz excitation: distribution of profile-wise NRMSEs over the full evaluation window $[0,8]$~s, with the mean and median indicated (left), and instantaneous NRMSE during rollout (right). The black curve shows the mean across the 500 test profiles; the gray band represents one population standard deviation above and below the mean, with the lower bound clipped at zero where necessary. Pink shading marks the extrapolation interval $[4,8]$~s. Errors are calculated from velocities in physical units, normalized by each profile's peak absolute reference velocity over the full evaluation window, and expressed as percentages. }
    \label{fig:deep_05Hz_error_dist}
\end{figure}

Figure~\ref{fig:deep_05Hz_error_dist} reports a mean NRMSE of 1.88\% and a median of 1.85\% over the full evaluation window. The histogram shows that most test profiles have comparatively low errors, with a smaller group producing larger discrepancies. Over the extrapolation interval alone, the mean and median NRMSEs are 2.59\% and 2.54\%, respectively (Table~\ref{tab:results_summary}). The mean instantaneous NRMSE varies during extrapolation, indicating changes in the spatial RMS discrepancy relative to each profile's fixed reference peak. Its temporal variation does not represent an error accumulated over time. The larger extrapolation-only NRMSE indicates that the later response has a higher RMS discrepancy relative to the event peak than the full-window average. Overall, this case demonstrates reproduction of the dominant long-period velocity response beyond the training interval, with larger aggregate discrepancies during the later response.

\subsection{Deep Profile with 3.5 Hz}
\label{subsec:deep_profile_3.5Hz}
\begin{figure}[htbp!]
    \centering
    \includegraphics[width=\linewidth]{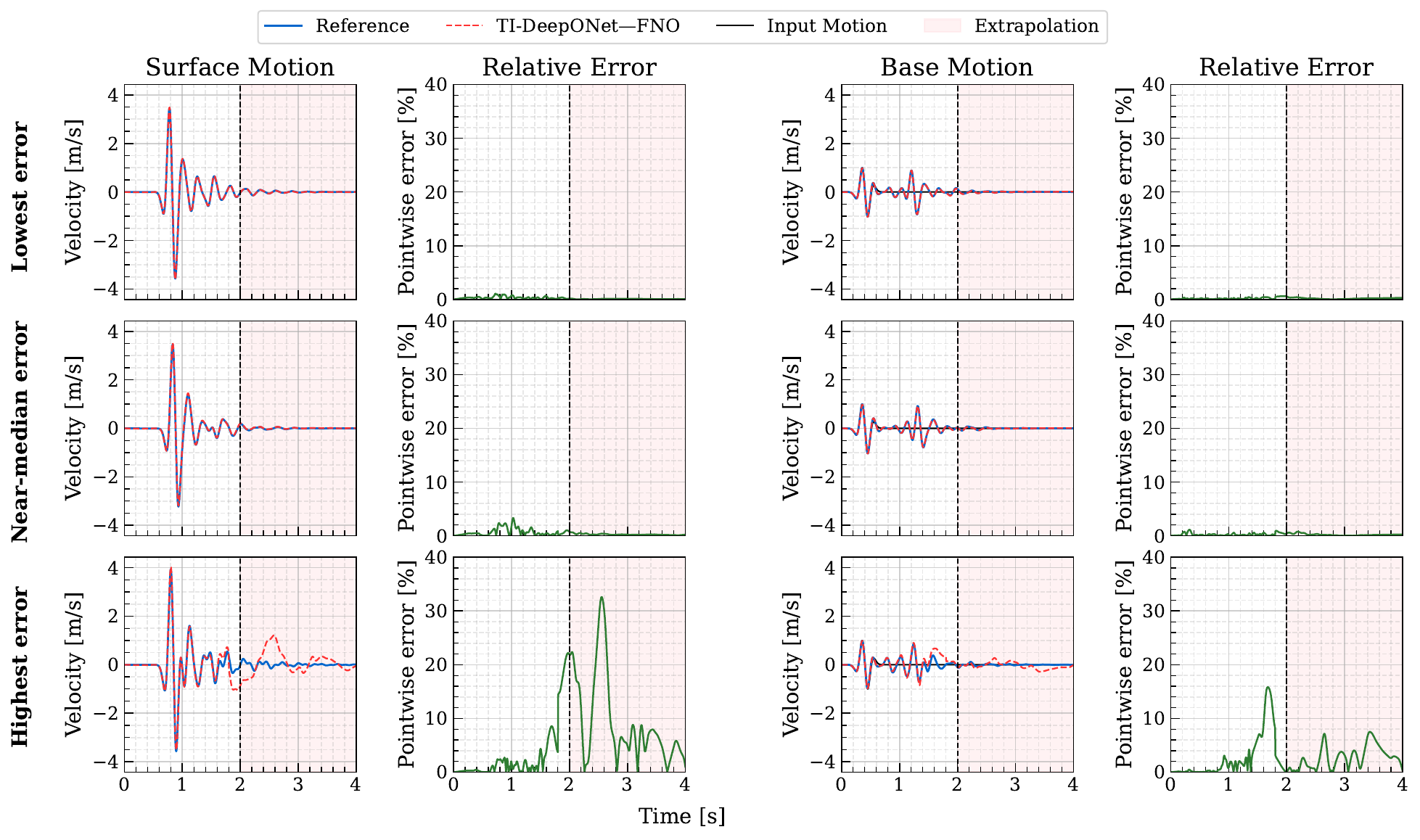}
    \caption{Deep profile with 3.5-Hz excitation: reference and TI-DeepONet--FNO surface and base velocity histories for profiles selected by their lowest, near-median, and highest NRMSEs over the full evaluation window. Pointwise errors at both stations are expressed as percentages of each profile's peak absolute reference surface velocity over the displayed evaluation window. The training interval is $[0,2)$~s, and shading marks the extrapolation interval $[2,4]$~s.}
    \label{fig:deep_35Hz_surface}
\end{figure}

\begin{figure}[htbp!]
    \centering
    \includegraphics[width=\linewidth]{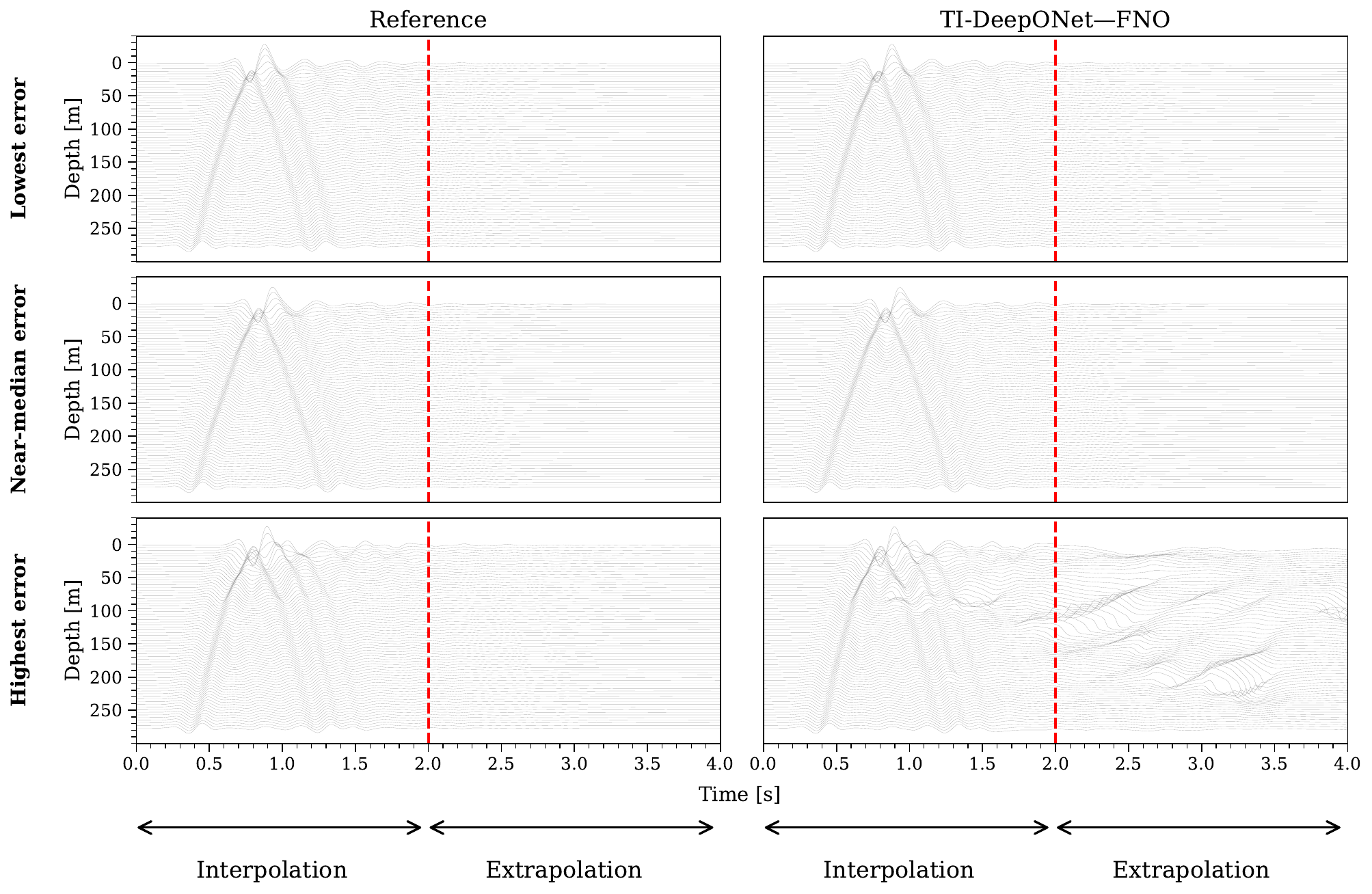}
    \caption{Deep profile with 3.5-Hz excitation: reference and TI-DeepONet--FNO depth--time velocity seismograms for the same profiles shown in Figure~\ref{fig:deep_35Hz_surface}. The training interval is $[0,2)$~s, and the extrapolation interval is $[2,4]$~s. Velocities are shown in physical units.}
    \label{fig:deep_35Hz_seismogram}
\end{figure}

The deep-profile case with 3.5-Hz excitation examines performance for a more rapidly varying response. Figures~\ref{fig:deep_35Hz_surface} and~\ref{fig:deep_35Hz_seismogram} present the surface and base velocity histories and the corresponding depth--time seismograms. Compared with the 0.5-Hz case, the higher excitation frequency reduces the characteristic wavelength and produces a more oscillatory response with a greater number of distinguishable reflected arrivals. These wavefield features can increase sensitivity to small phase or amplitude discrepancies as waves undergo repeated transmission and reflection at layer interfaces.

The predicted surface and base motions reproduce the principal arrivals for the selected lowest-error and near-median profiles, with close agreement in waveform amplitude and timing. The corresponding depth--time seismograms also capture the propagation trajectories and spatial structure of the dominant wave packets. These features remain continuous across the training--extrapolation boundary at $t=2$~s, indicating preservation of the principal temporal structure during extrapolation for these profiles. However, the selected highest-error profile exhibits substantially larger discrepancies, including distorted reflected arrivals and spurious late-time motion throughout the column. These comparisons illustrate variation in prediction quality across the selected profiles.

\begin{figure}[htbp!]
    \centering
    \includegraphics[width=\linewidth]{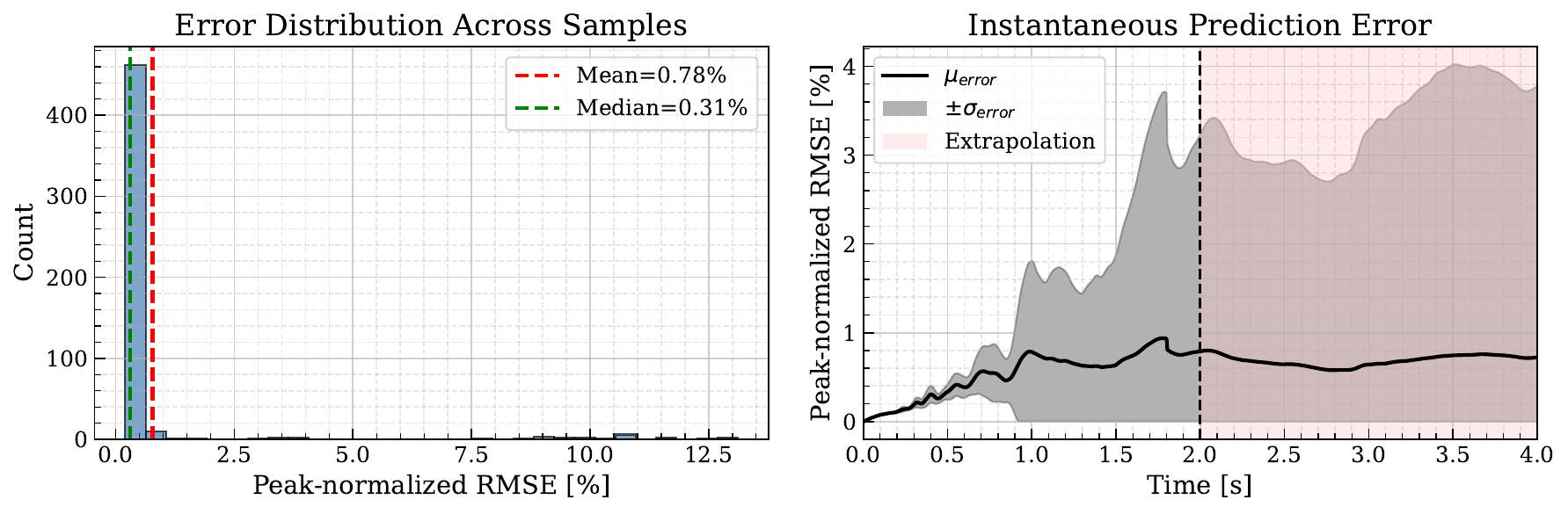}
    \caption{Deep profile with 3.5-Hz excitation: distribution of profile-wise NRMSEs over the full evaluation window $[0,4]$~s, with the mean and median indicated (left), and instantaneous NRMSE during rollout (right). The black curve shows the mean across the 500 test profiles; the gray band represents one population standard deviation above and below the mean, with the lower bound clipped at zero where necessary. Pink shading marks the extrapolation interval $[2,4]$~s. Errors are calculated from velocities in physical units, normalized by each profile's peak absolute reference velocity over the full evaluation window, and expressed as percentages.}
    \label{fig:deep_35Hz_error_dist}
\end{figure}

Figure~\ref{fig:deep_35Hz_error_dist} reports a mean NRMSE of 0.78\% and a median of 0.31\% over the full evaluation window. The distribution is strongly right-skewed: most profiles exhibit low errors, whereas a smaller group forms an extended upper tail. The selected highest-error profile has a reported NRMSE of approximately 13.11\% (Table~\ref{tab:results_summary}), illustrating the substantial discrepancies that can occur in individual configurations despite the low median error. Over the extrapolation interval alone, the mean and median NRMSEs are 0.73\% and 0.14\%, respectively. These values remain normalized by each profile's peak reference velocity over the full evaluation window. The standard-deviation band in Figure~\ref{fig:deep_35Hz_error_dist} describes the variability of instantaneous NRMSE across test profiles. Each profile uses a fixed reference amplitude, so temporal changes in its NRMSE reflect changes in the spatial RMS discrepancy relative to that amplitude. Where the standard deviation exceeds the mean, the lower bound of the plotted band is clipped at zero; this is a plotting convention rather than evidence of zero prediction error. Identifying which combinations of layer properties are associated with the larger errors requires further analysis.

\subsection{Shallow Profile with 3.5 Hz}
\label{subsec:shallow_profile_3.5Hz}
\begin{figure}[htbp!]
    \centering
    \includegraphics[width=\linewidth]{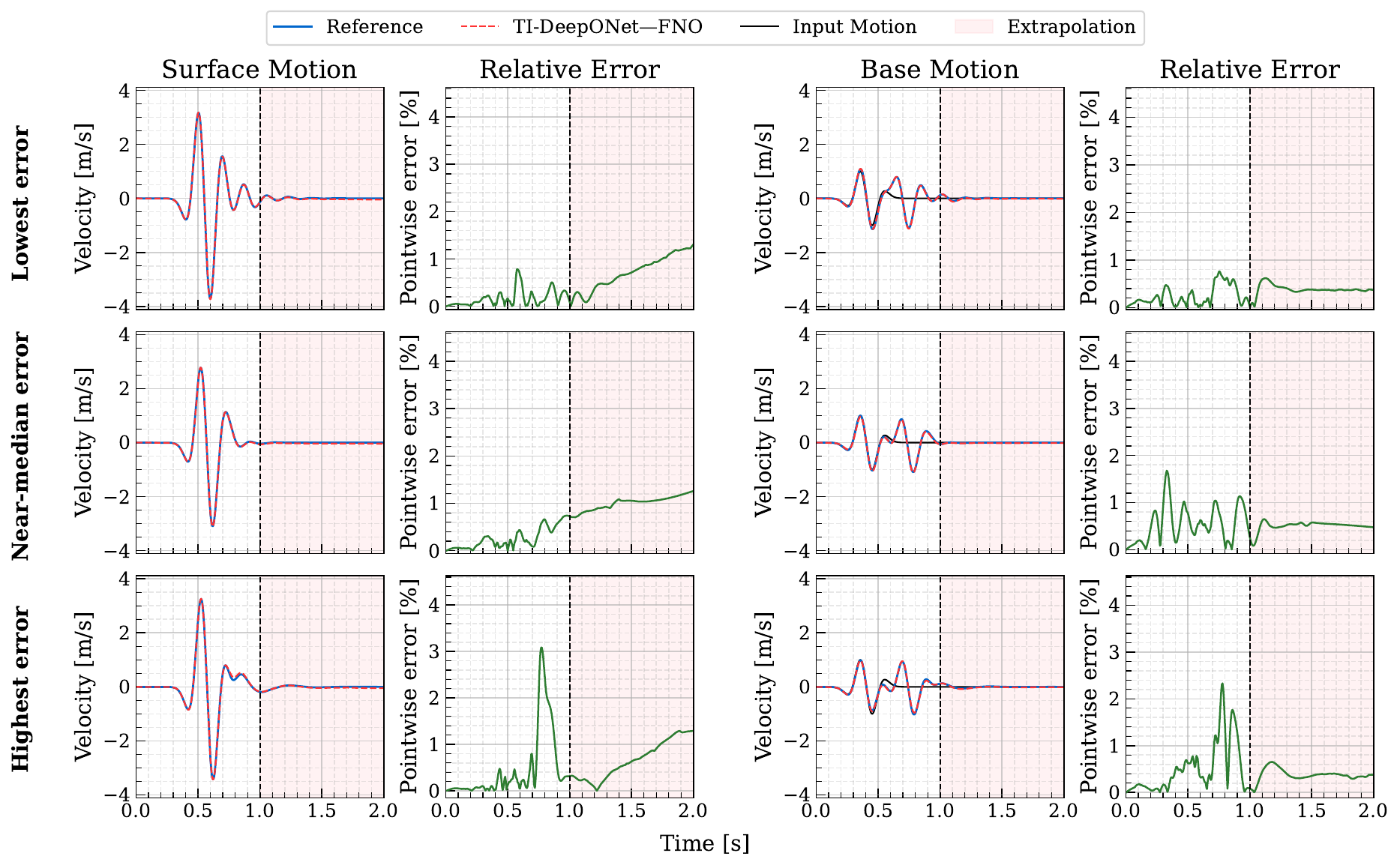}
    \caption{Shallow profile with 3.5-Hz excitation: reference and TI-DeepONet--FNO surface and base velocity histories for profiles selected by their lowest, near-median, and highest NRMSEs over the full evaluation window. Pointwise errors at both stations are expressed as percentages of each profile's peak absolute reference surface velocity over the displayed evaluation window. The training interval is $[0,1)$~s, and shading marks the extrapolation interval $[1,2]$~s.}
    \label{fig:shallow_35Hz_surface}
\end{figure}

\begin{figure}[htbp!]
    \centering
    \includegraphics[width=\linewidth]{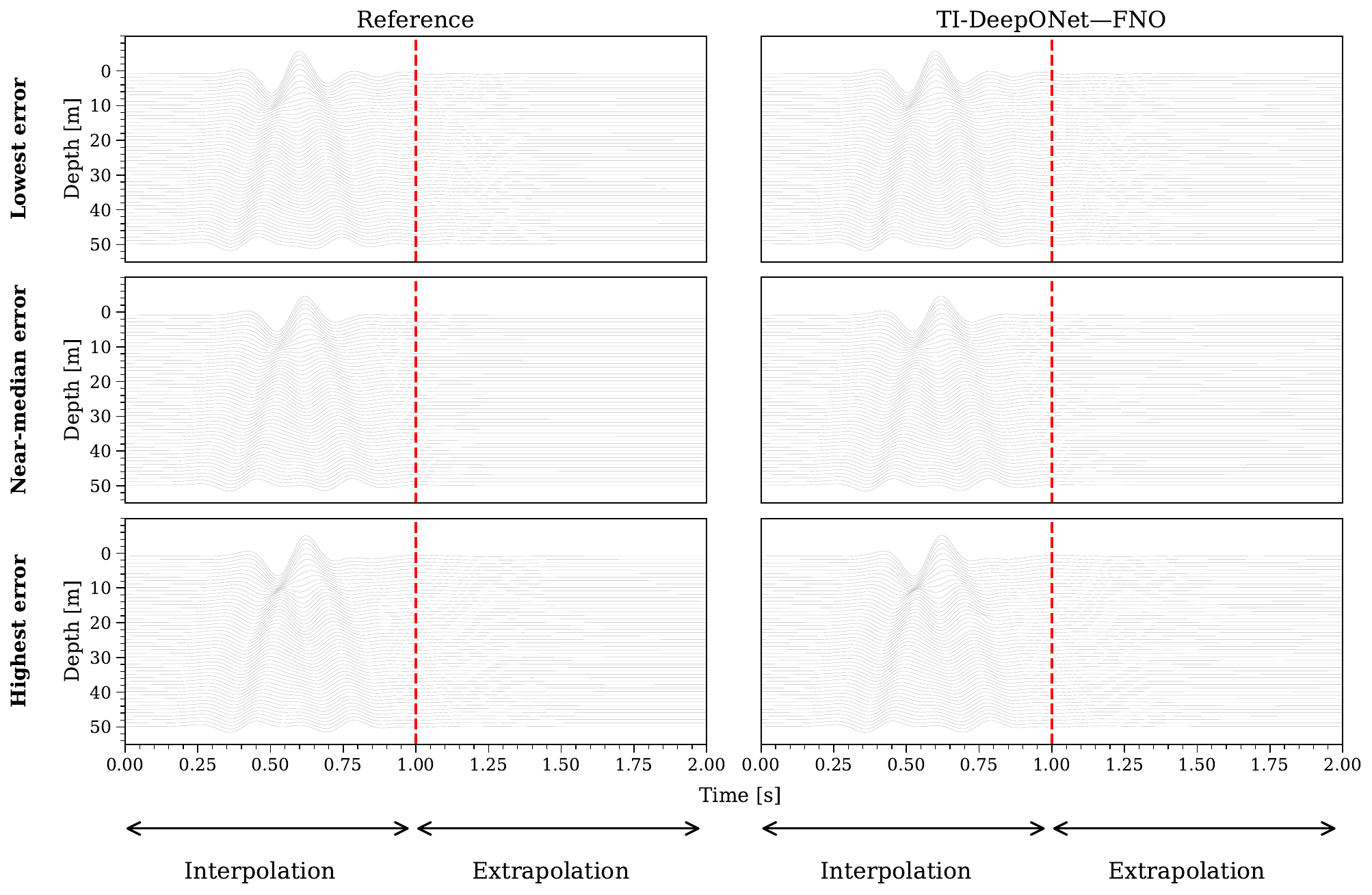}
    \caption{Shallow profile with 3.5-Hz excitation: reference and TI-DeepONet--FNO depth--time velocity seismograms for the same profiles shown in Figure~\ref{fig:shallow_35Hz_surface}. Dashed vertical lines mark the first extrapolation time level at $t=1$~s; the displayed evaluation window extends to $t=2$~s. Velocities are shown in physical units.}
    \label{fig:shallow_35Hz_seismogram}
\end{figure}

Figures~\ref{fig:shallow_35Hz_surface} and~\ref{fig:shallow_35Hz_seismogram} show the results for the 50-m soil profile subjected to 3.5-Hz excitation. Because of the short propagation distance, the incident and reflected waves arrive within a relatively compact time interval. The selected 1-s training window includes the dominant wave pulses, including the principal incident and reflected arrivals. The extrapolation interval primarily contains the decaying, low-amplitude portion of the response. Accordingly, this case evaluates prediction of the late-time velocity response after the main arrivals and provides a less stringent test of predicting distinct arrivals beyond the training interval.

\begin{figure}[htbp!]
    \centering
    \includegraphics[width=\linewidth]{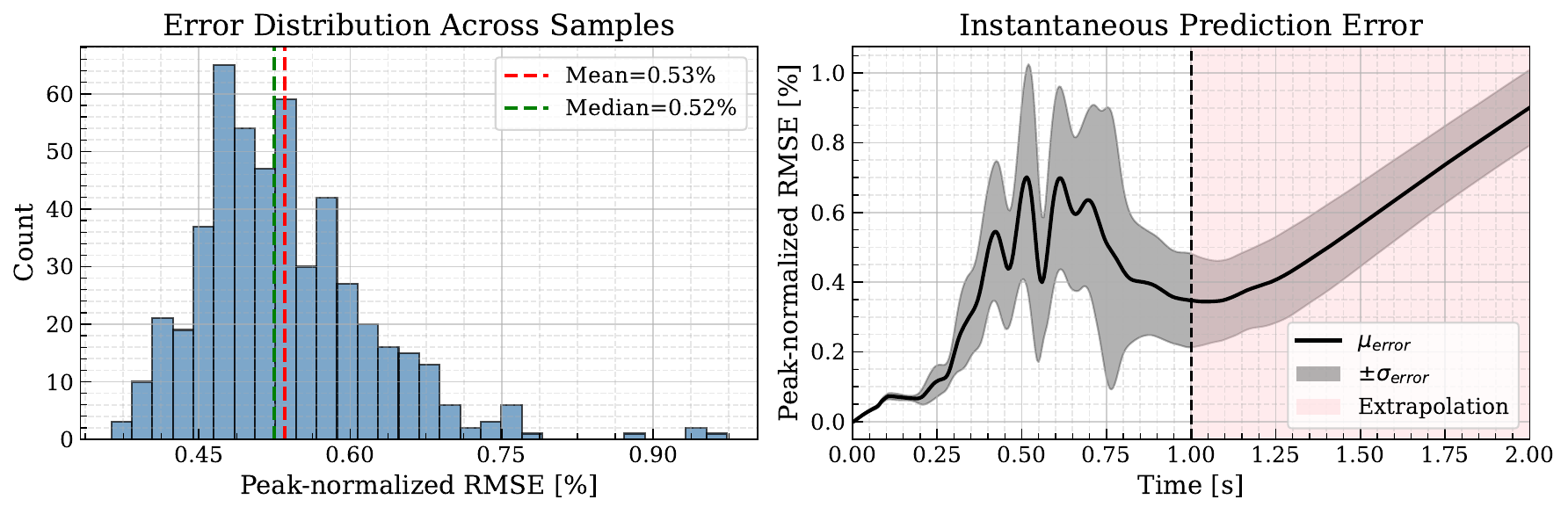}
    \caption{Shallow profile with 3.5-Hz excitation: distribution of profile-wise NRMSEs over the full evaluation window $[0,2]$~s, with the mean and median indicated (left), and instantaneous NRMSE during rollout (right). The black curve shows the mean across the 500 test profiles; the gray band represents one population standard deviation above and below the mean, with the lower bound clipped at zero where necessary. Pink shading marks the extrapolation interval $[1,2]$~s. Errors are calculated from velocities in physical units, normalized by each profile's peak absolute reference velocity over the full evaluation window, and expressed as percentages.}
    \label{fig:shallow_35Hz_error_dist}
\end{figure}

The proposed framework reproduces the dominant surface and base velocity histories for the three selected test profiles. The depth--time seismograms similarly show agreement in the arrival times, propagation trajectories, and depth-dependent amplitudes of the principal wave packets. Local discrepancies remain in the velocity histories, including during the high-amplitude response, where small timing or amplitude differences can produce appreciable pointwise errors. Figure~\ref{fig:shallow_35Hz_error_dist} complements these waveform comparisons by showing the distribution of profile-wise errors and the temporal variation in the mean instantaneous NRMSE.

The full-window NRMSE has a mean of 0.53\% and a median of 0.52\%. Over the extrapolation interval alone, the mean and median are 0.61\% and 0.60\%, respectively (Table~\ref{tab:results_summary}). These values indicate small RMS discrepancies relative to each profile's peak reference velocity, including during the lower-amplitude response following the dominant arrivals. The following 6-Hz case provides a more demanding test of predicting distinct reflected arrivals beyond the training interval: its shorter-duration excitation pulses improve arrival separation, allowing the selected shorter training window to include the principal surface-motion peak while leaving part of the reflected-wave sequence for extrapolation.

Preliminary single-run experiments with shorter training windows containing 100 and 150 time levels yielded median full-window NRMSEs of 45.3\% and 21.2\%, respectively, compared with 0.52\% for the reported 1-s window. All three experiments were evaluated over 0--2~s using the same 500 test profiles. The numbers of profiles whose peak predicted surface velocity exceeded twice the reference surface peak were 19, 219, and zero, respectively. Thus, the median error decreased with increasing window length, whereas the frequency of excessive amplitude growth did not vary monotonically. The shortest-window run used 200{,}000 training epochs, compared with 300{,}000 for the other two runs, and the number of available training pairs also varied with window length. These comparisons support the reported 1-s window under the tested settings without
establishing a general minimum training duration. Although the dominant surface peak was included for all profiles in the 150-time-level window, substantial extrapolation errors persisted, indicating that inclusion of this peak alone was insufficient for reliable prediction.

\subsection{Shallow Profile with 6 Hz}
\label{subsec:shallow_profile_6Hz}
\begin{figure}[htbp!]
    \centering
    \includegraphics[width=\linewidth]{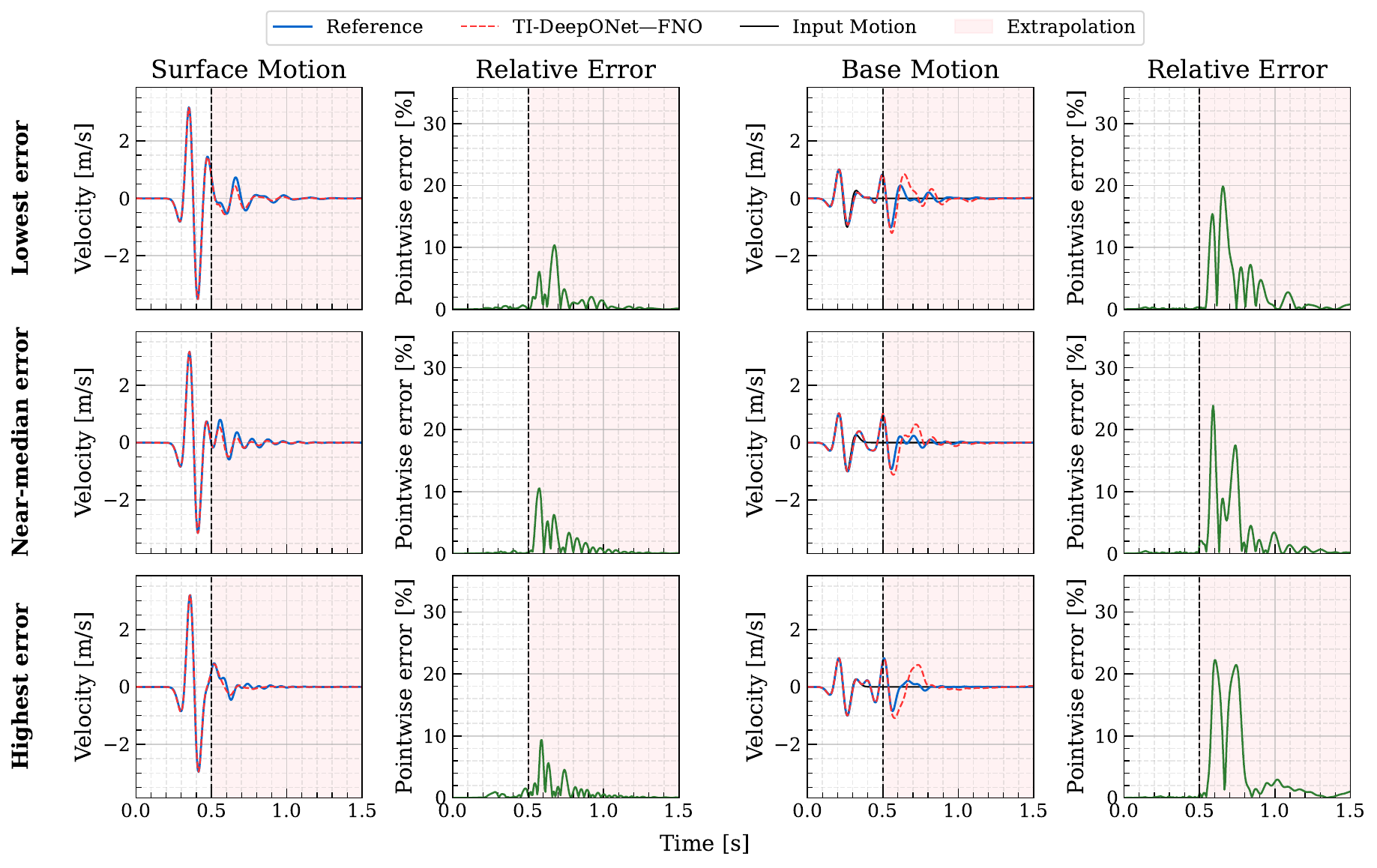}
    \caption{Shallow profile with 6-Hz excitation: reference and TI-DeepONet--FNO surface and base velocity histories for profiles selected by their lowest, near-median, and highest NRMSEs over the full evaluation window. Pointwise errors at both stations are expressed as percentages of each profile's peak absolute reference surface velocity over the displayed evaluation window. The training interval is $[0,0.5)$~s, and shading marks the extrapolation interval $[0.5,1.5]$~s.}
    \label{fig:shallow_6Hz_surface}
\end{figure}

\begin{figure}[htbp!]
    \centering
    \includegraphics[width=\linewidth]{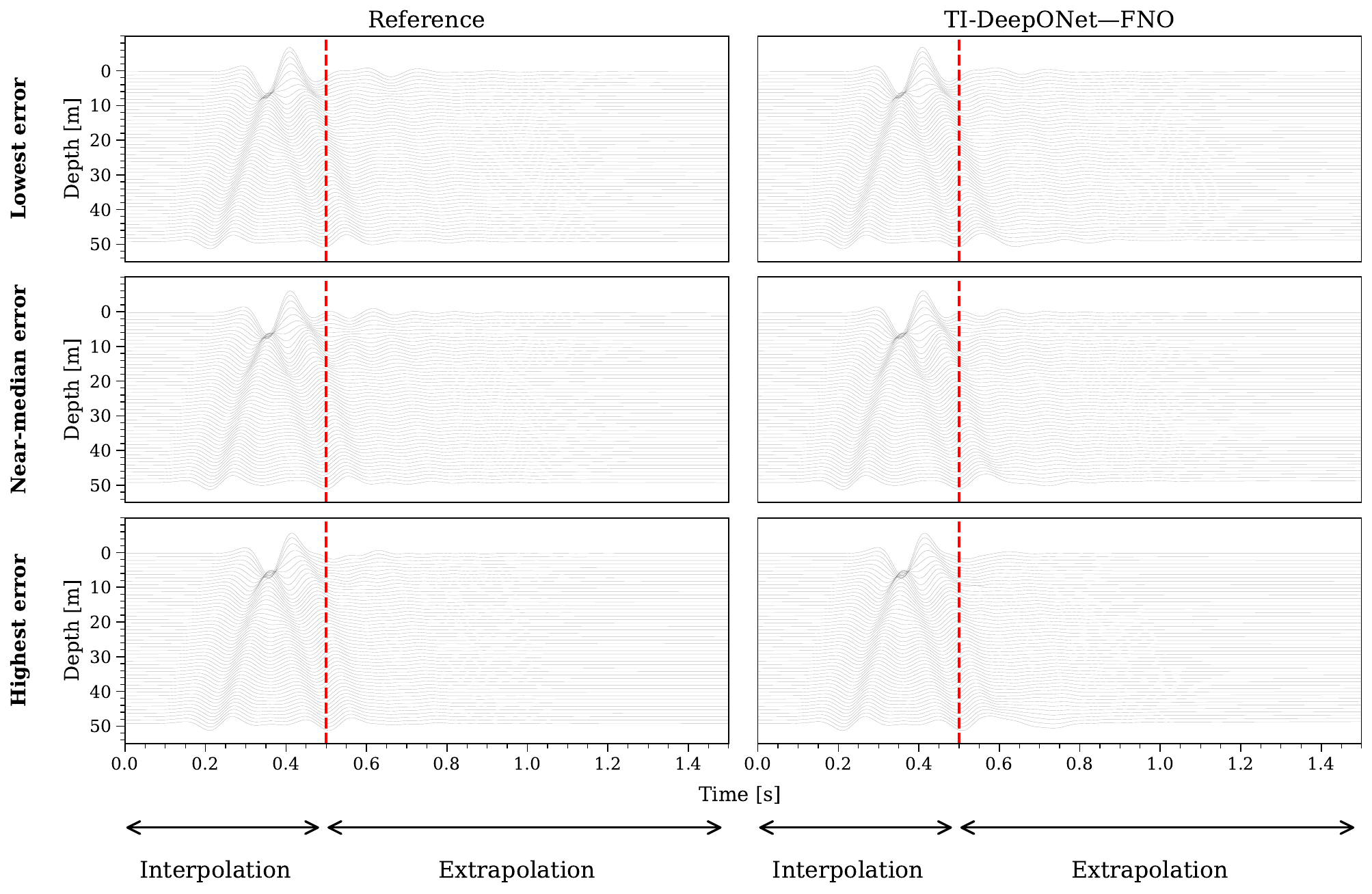}
    \caption{Shallow profile with 6-Hz excitation: reference and TI-DeepONet--FNO depth--time velocity seismograms for the same profiles shown in Figure~\ref{fig:shallow_6Hz_surface}. The training interval is $[0,0.5)$~s, and the extrapolation interval is $[0.5,1.5]$~s. Velocities are shown in physical units.}
    \label{fig:shallow_6Hz_seismogram}
\end{figure}

For the shallow profile, increasing the excitation frequency from 3.5 to 6~Hz reduces the characteristic wavelengths and shortens the input pulse, making individual wave arrivals easier to distinguish. Figure~\ref{fig:shallow_6Hz_surface} shows that TI-DeepONet--FNO captures the dominant surface pulse and the main base-motion arrivals in all three selected profiles. The depth--time seismograms in Figure~\ref{fig:shallow_6Hz_seismogram} also reproduce the principal propagation patterns beyond the first extrapolation time level at $t=0.5$~s. Agreement is closer at the surface than at the base, where differences in the amplitudes and timing of subsequent arrivals are more apparent. Thus, the shorter training window allows distinct arrivals to be evaluated during extrapolation while exposing discrepancies that are less evident in the dominant surface response.

\begin{figure}[htbp!]
    \centering
    \includegraphics[width=\linewidth]{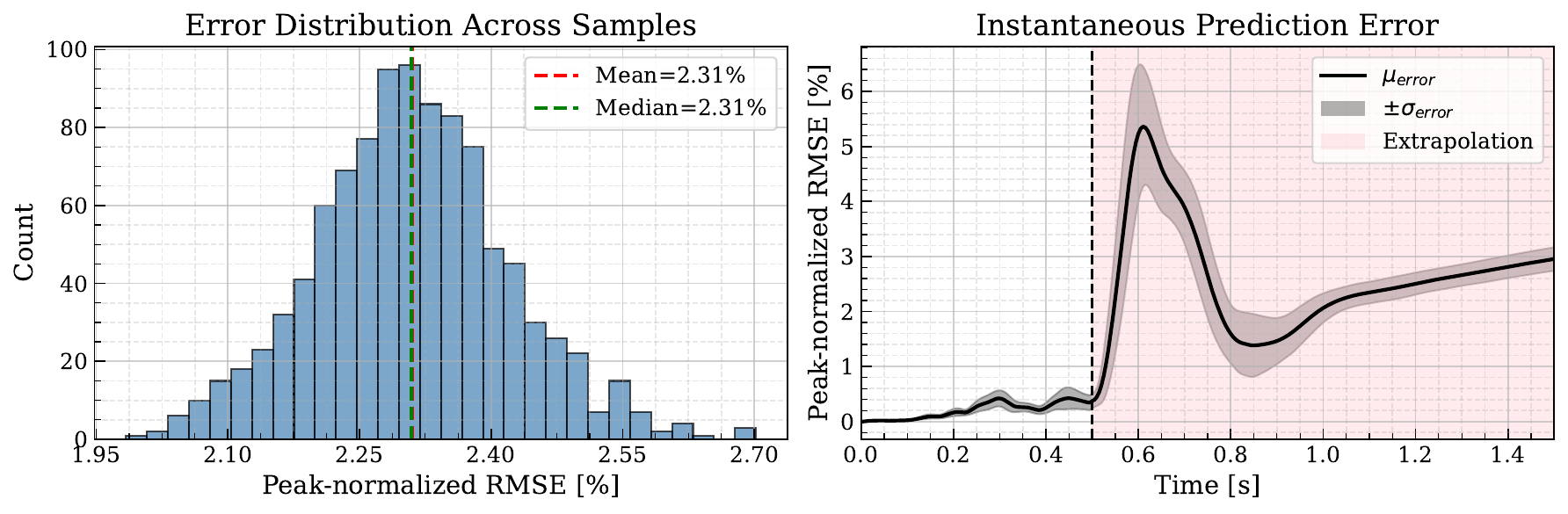}
    \caption{Shallow profile with 6-Hz excitation: distribution of profile-wise NRMSEs over the full evaluation window $[0,1.5]$~s, with the mean and median indicated (left), and instantaneous NRMSE during rollout (right). The black curve shows the mean across the 1000 test profiles; the gray band represents one population standard deviation above and below the mean, with the lower bound clipped at zero where necessary. Pink shading marks the extrapolation interval $[0.5,1.5]$~s. Errors are calculated from velocities in physical units, normalized by each profile's peak absolute reference velocity over the full evaluation window, and expressed as percentages.}
    \label{fig:shallow_6Hz_error_dist}
\end{figure}

The full-window NRMSE has a mean and median of 2.31\% (Figure~\ref{fig:shallow_6Hz_error_dist}). Both values exceed those of the shallow 3.5-Hz case. Over the extrapolation interval alone, both the mean and median are 2.82\% (Table~\ref{tab:results_summary}). The mean instantaneous NRMSE rises after the training--extrapolation boundary, then decreases before increasing again toward the end of the evaluation window. Because each profile uses a fixed reference peak, this non-monotonic behavior reflects changes in the spatial RMS discrepancy relative to that peak. The station histories show more pronounced base-motion discrepancies around the post-training arrivals, with smaller differences persisting later in the response.

Across all four cases, the TI-DeepONet--FNO framework achieves mean NRMSEs ranging from 0.53\% to 2.31\% over the full evaluation windows and from 0.61\% to 2.82\% over the extrapolation intervals, with additional prediction durations up to twice the training duration. Accuracy varies across cases: the shallow 6-Hz case has higher mean errors than the shallow 3.5-Hz case, whereas the deep 3.5-Hz case has lower mean errors than the deep 0.5-Hz case. Because the training and extrapolation durations also differ, these comparisons do not isolate the effect of excitation frequency. The selected velocity histories and seismograms demonstrate reproduction of principal arrivals and propagation patterns beyond the training interval, alongside localized discrepancies and substantial errors in some profiles. Together, these results support temporal velocity prediction across the tested layered configurations under continuing reference-stress conditioning, while identifying reflected arrivals and late-time motion as features requiring further improvement.

\section{Practical Applicability, Limitations, and Future Directions}
\label{sec:applicability}
The layered elastic configuration considered in Section~\ref{sec:problem_statement} provides a physically interpretable setting for studying seismic response governed by impedance contrasts, reflection, transmission, and wave interference. These mechanisms control travel times and frequency-dependent amplification, making elastic models useful for interpreting small-strain wave propagation, examining stratigraphic effects, and establishing benchmarks for numerical and learned prediction methods \cite{thomson1950Transmission,haskell1953Dispersion, kennett1979Seismic, thompson2012Taxonomy, regnier2016International}. The elastic assumption isolates these propagation mechanisms from strain-dependent stiffness degradation and material dissipation; however, small-strain soil response can still exhibit damping, whose representation may be needed to reproduce resonance amplitudes and response duration quantitatively \cite{asimaki2017SeismoSoil, lanzo1999Effect, tao2019Insights, zalachoris2015Evaluation}. Within this idealized setting, the results in Section~\ref{sec:results} demonstrate that the jointly trained temporal predictor and spatial corrector reproduce principal velocity arrivals and depth-dependent response patterns across variable layered profiles, including at times beyond the training interval. As defined by the coupled operator in Eq.~\eqref{eq:coupled_operator}, this capability relies on reference stress fields supplied throughout rollout and therefore represents conditional velocity prediction rather than a standalone simulation of the coupled velocity--stress system.

The results in Section~\ref{sec:results} motivate a systematic examination of how excitation frequency, training-window duration, and extrapolation duration influence prediction accuracy. Controlled comparisons should account for changes in the number of training pairs and training budget when varying these quantities. The preliminary shallow 3.5-Hz experiments described in Section~\ref{subsec:shallow_profile_3.5Hz} provide a starting point for examining how much of the incident and reflected response should be represented during training to achieve a specified prediction accuracy. These experiments also indicate that including the dominant surface peak alone does not ensure reliable extrapolation. The high-error tail observed in the deep 3.5-Hz case further motivates investigating associations between prediction errors and impedance contrasts, layer travel times, and wavelength-to-thickness ratios. Such analyses would help characterize the conditions under which the demonstrated accuracy persists and identify where broader training coverage or targeted model refinements could improve prediction performance.

Further development can broaden both the physical conditions represented by the framework and its prediction capabilities. Evaluation with inputs spanning broader frequency bands and with recorded earthquake motions would extend the present DOG3-wavelet tests to more varied excitation histories. Incorporating viscoelastic or nonlinear constitutive behavior and higher-dimensional wave propagation would allow material dissipation, strain-dependent response, and more complex spatial interactions to be examined through appropriately designed training datasets and validation studies. A complementary development is to couple velocity prediction with a learned or physics-based stress update, enabling investigation of autonomous velocity--stress evolution and the control of errors arising from feedback between the two fields. At the architectural level, the current single FNO correction stage could be extended to multiple successive residual corrections. The value of additional stages should be established through controlled comparisons with the existing jointly trained framework, considering prediction accuracy, rollout stability, and computational cost.

\section{Conclusions}
\label{sec:conclusions}
This study developed a jointly trained TI-DeepONet--FNO framework for stress-conditioned temporal prediction of seismic velocity fields in layered elastic media. The TI-DeepONet advances the velocity through an embedded fourth-order Runge--Kutta scheme, while the FNO applies a learned spatial residual correction to the preliminary prediction. Following initialization with reference velocity fields at the first two time levels, the framework predicts velocity autoregressively, with reference stress fields and prescribed excitation supplied throughout rollout, including beyond the training interval.

The framework was evaluated on four cases of one-dimensional SH-wave propagation through four-layer soil columns with total depths of 50 and 280~m and selected excitation frequencies between 0.5 and 6~Hz. Across test profiles with varying layer thicknesses and shear-wave velocities, predictions extended beyond the training interval by up to twice its duration. The selected velocity histories and depth--time seismograms showed agreement with the reference solutions in the principal arrivals and spatial propagation patterns, including reflected arrivals beyond the training interval. Mean peak-normalized RMSEs ranged from 0.53\% to 2.31\% over the full evaluation windows and from 0.61\% to 2.82\% over the extrapolation intervals. These aggregate results coexist with localized waveform discrepancies and substantially larger errors in some profiles, demonstrating variation in prediction performance across the tested configurations.

These findings demonstrate that a jointly trained combination of learned temporal integration and spatial residual correction can predict velocity evolution beyond the training interval under the tested conditions. The framework provides a basis for investigating temporal generalization in oscillatory, interface-influenced wave propagation under reference-stress conditioning. Extending this formulation to jointly advance velocity and stress is a direct next step toward autonomous simulation of the coupled wavefield.

%Goes only for the camera ready version
\section*{Acknowledgments}
This research was partly supported by the National Science Foundation (NSF) under Grant Nos.~2438193 and 2436738. The University of Utah authors acknowledge support from the Department of Civil and Environmental Engineering at the University of Utah. The authors acknowledge computing support provided by the Advanced Research Computing at Hopkins (ARCH) core facility at Johns Hopkins University and the Rockfish cluster. The ARCH core facility (\url{https://rockfish.jhu.edu}) is supported by NSF Grant No.~OAC1920103. The support and resources from the Center for High Performance Computing at the University of Utah are gratefully acknowledged.

\section*{Data and code availability}
All data needed to evaluate the conclusions in the paper are presented in the paper. All code and data accompanying this manuscript will become publicly available at \url{https://github.com/Centrum-IntelliPhysics/NeuralOperators_for_WavePropagation} upon publication of the paper.

\section*{Author contributions}
\textbf{Dibyajyoti Nayak:} Conceptualization, Methodology, Software, Validation, Formal analysis, Investigation, Data Curation, Writing -- Original Draft, Writing -- Review \& Editing, Visualization;
\textbf{Yuze Pu:} Conceptualization, Methodology, Software, Validation, Formal analysis, Investigation, Data Curation, Writing -- Original Draft, Writing -- Review \& Editing, Visualization;
\textbf{Kami Mohammadi:} Conceptualization, Methodology, Validation, Formal analysis, Resources, Writing -- Review \& Editing, Visualization, Supervision, Project administration, Funding acquisition;
\textbf{Somdatta Goswami:} Conceptualization, Methodology, Validation, Formal analysis, Resources, Writing -- Review \& Editing, Visualization, Supervision, Project administration, Funding acquisition

\bibliographystyle{elsarticle-num} 
\bibliography{references.bib}

\newpage
\appendix
% \section{Noise level}

% For the shallow-profile case using a 6 Hz DOG3 wavelet, we fix the number of training time steps at 100 and vary the noise level to evaluate its influence. The noise level ranges from \(2\times10^{-3}\) to \(4\times10^{-3}\) in increments of \(5\times10^{-4}\).

%Shallow 3.5 check for ica100, inference all check for ica100
\section{Computational Costs}
\label{sec:computational_costs}
\begin{table}[htbp!]
\renewcommand{\arraystretch}{1.12}
\centering
\caption{Computational costs of training and inference for all cases. $N_{train}$ denotes the number of profiles contributing training pairs over $[0, T_{tr})$, and $N_{test}$ the number of profiles evaluated during rollout over $[0, T]$; inference times are measured over these $N_{test}$ profiles. Training for all the cases was carried out on an NVIDIA A100 GPU with 80 GB memory, paired with an Intel Xeon Gold Ice Lake CPU.}
\begin{tabular}{|c|c|c|c|c|c|c|}
\hline
Problem & $\#$Params & \makecell{Batch \\ size} & $N_{train}$ & \makecell{Training time \\ (iter/sec)} & $N_{test}$ & \makecell{Inference time \\ (sec)} \\
\hline
Deep 0.5 Hz & 1,368,509 & 64 & 500 & 1.85 & 500 & 44.11 \\
\hline
Deep 3.5 Hz & 1,630,653 & 64 & 500 & 3.62 & 500 & 30.62 \\
\hline
Shallow 3.5 Hz & 1,589,693 & 64 & 500 & 1.97 & 500 & 10.46 \\
\hline
Shallow 6 Hz & 1,589,693 & 64 & 1000 & 1.88 & 1000 & 12.03 \\
\hline
\end{tabular}
\label{tab:training-inference-times}
\end{table}

\section{Architecture Details}
\label{sec:architecture_details}
\begin{table}[htbp!]
    \renewcommand{\arraystretch}{1.12}
    \centering
    \caption{Architecture details of TI-DeepONet coupled with FNO correction network for all cases.}
    \begin{tabular}{|c|c|c|c|c|c|c|c|c|}
        \hline
        \multirow{3}{*}{Problem} & \multicolumn{3}{c|}{Branch/Trunk Net} & \multicolumn{4}{c|}{FNO Correction Network} & \multirow{3}{*}{Epochs} \\
        \cline{2-8}
        & Layers & \makecell{Hidden \\ Dim} & \makecell{Latent \\ Size} & \makecell{In/Out \\ Channels} & \makecell{Fourier \\ Blocks} & Modes & Width & \\
        \hline
        \makecell[c]{Deep \\ 0.5 Hz} & 5 & 128 & 100 & 3/1 & 4 & 24 & 64 & $3\times 10^5$ \\
        \hline
        \makecell[c]{Deep \\ 3.5 Hz} & 5 & 128 & 100 & 3/1 & 4 & 32 & 64 & $3\times 10^5$ \\
        \hline
        \makecell[c]{Shallow \\ 3.5 Hz} & 5 & 128 & 100 & 3/1 & 4 & 32 & 64 & $3\times 10^5$ \\
        \hline
        \makecell[c]{Shallow \\ 6 Hz} & 5 & 128 & 100 & 3/1 & 4 & 32 & 64 & $3\times 10^5$ \\
        \hline
    \end{tabular}%
    \label{tab:architecture-details}
\end{table}

\end{document}